\documentclass[journal]{IEEEtran}

\ifCLASSOPTIONcompsoc
  \usepackage[nocompress]{cite}
\else
  \usepackage{cite}
\fi

\usepackage{makecell}
\usepackage{ragged2e}
\usepackage{algorithm}
\usepackage{algorithmic}
\usepackage{booktabs}
\usepackage{multirow}
\usepackage{graphicx}
\usepackage{subcaption}
\usepackage{amsmath,amssymb,bm}
\usepackage{url}
\usepackage{newtxtext,newtxmath}
\usepackage{hyperref}   
\usepackage{cleveref}
\usepackage{balance}

\crefname{figure}{Fig.}{Figs.}
\Crefname{figure}{Fig.}{Figs.}

\crefname{section}{Sec.}{Secs.}
\Crefname{section}{Sec.}{Secs.}

\crefname{table}{Tab.}{Tabs.}
\Crefname{table}{Tab.}{Tabs.}

\crefname{equation}{Eq.}{Eqs.}
\Crefname{equation}{Eq.}{Eqs.}
\begin{document}
\title{ER-Pose: Rethinking Keypoint-Driven Representation Learning for Real-Time Human Pose Estimation}

\author{Nanjun Li,
        Pinqi Cheng,
        Zean Liu,
        Minghe Tian,
        Xuanyin Wang*
        

}

\IEEEtitleabstractindextext{%
\begin{abstract}
\justifying
Single-stage multi-person pose estimation aims to jointly perform human localization and keypoint prediction within a unified framework, offering advantages in inference efficiency and architectural simplicity. Consequently, multi-scale real-time detection architectures, such as YOLO-like models, are widely adopted for real-time pose estimation. However, these approaches typically inherit a box-driven modeling paradigm from object detection, in which pose estimation is implicitly constrained by bounding-box supervision during training. This formulation introduces biases in sample assignment and feature representation, resulting in task misalignment and ultimately limiting pose estimation accuracy.
In this work, we revisit box-driven single-stage pose estimation from a keypoint-driven perspective and identify semantic conflicts among parallel objectives as a key source of performance degradation. To address this issue, we propose a keypoint-driven learning paradigm that elevates pose estimation to a primary prediction objective. Specifically, we remove bounding-box prediction and redesign the prediction head to better accommodate the high-dimensional structured representations for pose estimation. We further introduce a keypoint-driven dynamic sample assignment strategy to align training objectives with pose evaluation metrics, enabling dense supervision during training and efficient NMS-free inference. In addition, we propose a smooth OKS-based loss function to stabilize optimization in regression-based pose estimation.
Based on these designs, we develop a single-stage multi-person pose estimation framework, termed ER-Pose. On MS COCO and CrowdPose, ER-Pose-n achieves AP improvements of 3.2/6.7 without pre-training and 7.4/4.9 with pre-training respectively compared with the baseline YOLO-Pose. These improvements are achieved with fewer parameters and higher inference efficiency.
The source code will be made publicly available upon acceptance.
\end{abstract}

}

\maketitle

\IEEEdisplaynontitleabstractindextext

%
\IEEEpeerreviewmaketitle

\section{Introduction}
\label{sec:introduction}
\IEEEPARstart{R}{eal-time} multi-person pose estimation aims to efficiently and accurately predict human keypoints and their underlying spatial structure from a single image.
It serves as a fundamental technology in applications such as intelligent surveillance, human–computer interaction, and motion analysis
\cite{zheng2023deep,dang2019deep}.
As practical systems increasingly demand higher inference efficiency, deployment robustness and real-time responsiveness,
achieving efficient inference while maintaining high accuracy
has become one of the central challenges in 2D multi-person pose estimation.

In recent years, single-stage multi-person pose estimation methods have gained considerable attention due to their end-to-end inference paradigm and high efficiency.
By jointly performing human instance localization and keypoint prediction within a unified network,
these approaches substantially simplify the inference pipeline and demonstrate strong applicability in real-time scenarios.

Motivated by the success of real-time object detection, particularly YOLO-based architectures,
multi-scale representation strategies have been increasingly adopted for multi-person pose estimation \cite{redmon2016you,maji2022yolo,lu2024rtmo,mcnally2022rethinking}.
Such methods model human instances in multi-scale feature spaces and complete instance localization and keypoint prediction in a single forward pass,
achieving competitive real-time performance and practical deployment potential.

However, from the representation learning perspective,
these multi-scale designs are primarily tailored for human instance detection and localization.
Pose estimation is typically conditioned on detection outputs,
with keypoint prediction treated as a secondary objective within the overall modeling framework.
Under such a box-driven paradigm,
sample assignment strategies, multi-scale feature organization, and optimization objectives
are systematically structured around instance detection,
resulting in an inherent mismatch with the high-dimensional structured regression objective intrinsic to pose estimation.

Moreover, these approaches commonly rely on non-maximum suppression (NMS) based on instance localization during inference.
In crowded or heavily occluded scenarios,
this procedure may suppress candidates with superior pose quality and lead to identity confusion among instances,
thereby limiting the overall performance ceiling.
We refer to this detection-centered modeling paradigm, which organizes the pose prediction pipeline around bounding box localization, as the box-driven paradigm.

It is worth noting that, in object detection,
regression-based multi-scale representations have long been considered limited in accuracy due to sparse supervision and insufficient structural expressiveness.
To overcome these limitations, numerous studies have improved detection performance by increasing model capacity, redesigning network modules, or introducing more sophisticated post-processing strategies
\cite{bochkovskiy2020yolov4,ge2021yolox,wang2023yolov7,wang2024yolov10,khanam2024yolov11}.
However, directly transferring this line of optimization to multi-person pose estimation has not yielded comparable gains.
Existing single-stage pose estimation methods built upon multi-scale detection representations
often exhibit diminishing returns in pose accuracy despite increasing model scale or architectural complexity
\cite{ultralytics_yolov5,maji2022yolo,ultralytics_yolov8_pose,khanam2024yolov11}.

We argue that this phenomenon does not arise from inherent limitations of regression-based pose representations,
but rather from the systematic constraints imposed by the box-driven modeling assumption on the pose regression task.
When pose estimation is not formulated as a primary optimization objective,
its structured regression nature cannot be fully exploited.

Based on the above analysis,
we revisit the box-driven multi-scale single-stage pose estimation paradigm
and propose a keypoint-driven regression modeling framework that explicitly prioritizes pose prediction quality.
Building upon existing multi-scale detection-based pose estimators,
we systematically reformulate the framework from three perspectives: network architecture, sample assignment mechanism, and optimization objective.
While preserving the advantages of multi-scale representations,
our design removes the persistent interference of detection semantics on pose feature learning and regression optimization.
Specifically, we introduce Efficient-Regression Pose (ER-Pose),
a YOLO-like multi-scale single-stage real-time multi-person pose estimation network.
By migrating the multi-head assignment mechanism to the pose estimation task,
ER-Pose eliminates the need for any post-processing during inference.
During training, we analyze multi-task semantic bias and mathematically model task misalignment in multi-person pose estimation.
Based on this formulation, we design a keypoint-driven assignment metric that mitigates inference-time task bias and consistently improves prediction accuracy.
Furthermore, we reformulate keypoint regression variables by explicitly incorporating keypoint scale and uncertainty,
and propose the Smooth-OKS loss to globally optimize pose training.

\begin{figure*}[!h]
  \centering
  \begin{subfigure}{0.45\textwidth}
      \centering
      \includegraphics[width=\linewidth]{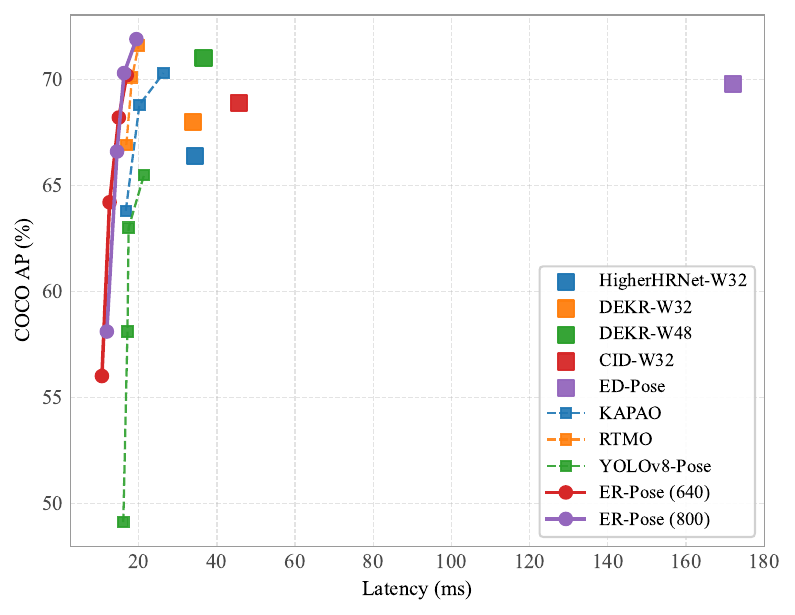}
      \label{figure:times vs ap}
  \end{subfigure}
  \begin{subfigure}{0.45\textwidth}
      \centering
      \includegraphics[width=\linewidth]{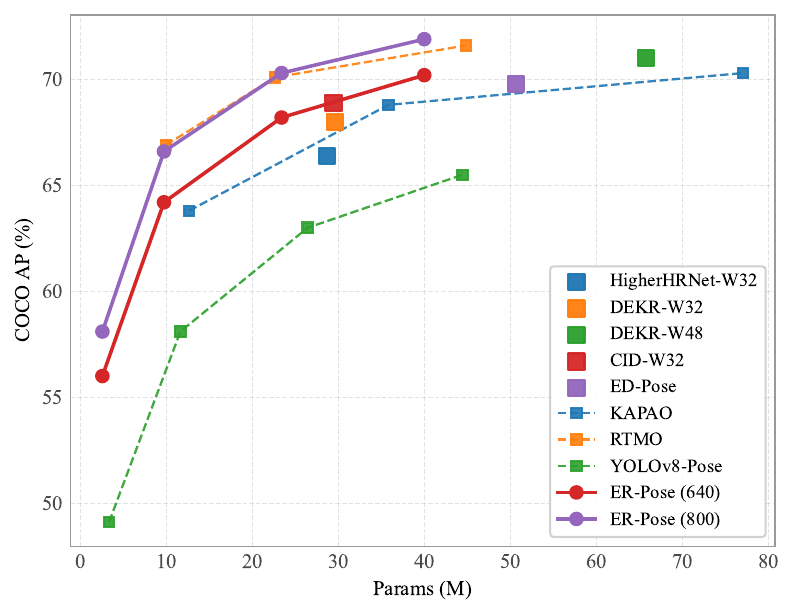}
      \label{figure:params vs ap}
  \end{subfigure}
\caption{Performance comparison between ER-Pose and representative human pose estimation methods on 
the COCO test-dev set.
    (a) Accuracy vs. inference time.
    (b) Accuracy vs. number of parameters.
    ER-Pose demonstrates superior inference efficiency while maintaining competitive accuracy.}
  \label{figure:performance}
\end{figure*}

Extensive experiments were conducted on COCO Keypoint \cite{lin2014microsoft}, CrowdPose \cite{li2019crowdpose}, and OCHuman \cite{zhang2019pose2seg}.
The results demonstrate that ER-Pose achieves performance competitive with state-of-the-art methods while using fewer parameters and attaining higher inference speed.
As shown in Fig.~\ref{figure:performance},
ER-Pose achieves state-of-the-art pose estimation accuracy on COCO while maintaining faster inference speed.
Compared with the baseline YOLO-v8-n, ER-Pose-n reduces parameter count by 23.3\%
and achieves accuracy improvements of +6.7 AP while operating at lower latency.
Moreover, ER-Pose exhibits enhanced robustness in challenging scenarios such as occlusion and motion blur.

Notably, ER-Pose relies solely on end-to-end offset regression without any post-processing operations,
yet attains state-of-the-art performance.
These results suggest that, under appropriate architectural design,
regression-based approaches can fully exploit their structured prediction capability
while retaining their inherent efficiency advantages.

The main contributions of this work are summarized as follows:
\begin{itemize}
    \item We provide a systematic analysis of the semantic bias induced by box-driven modeling and establish a keypoint-driven regression paradigm termed Efficient Regression Pose (ER-Pose) for single-stage multi-person pose estimation.
    \item We design a keypoint-driven dynamic sample assignment strategy, termed keypoint-driven Assignment, which leverages a joint OKS–confidence metric to enhance alignment between training objectives and inference behavior.
    \item We propose a smooth loss function for regression-based pose estimation, termed Smooth-OKS (SOKS), which unifies the properties of Gaussian and Laplace distributions to enhance gradient continuity and stabilize optimization.
    \item Extensive experiments on MS COCO, CrowdPose, and OCHuman demonstrate that ER-Pose achieves consistent accuracy improvements while maintaining NMS-free inference and high computational efficiency, validating the effectiveness of the proposed keypoint-driven paradigm.
\end{itemize}

\section{Related Work}
\subsection{Multi-Person Pose Estimation Methods}

Multi-person pose estimation approaches are generally categorized into top-down, bottom-up, and single-stage methods based on their detection pipeline \cite{dang2019deep,zheng2023deep}.  
Among these, top-down and bottom-up approaches are collectively termed two-stage methods.

Top-down methods \cite{sun2019deep,yu2021lite,yang2021transpose,yuan2021hrformer,cheng2020higherhrnet,
xiao2018simple,newell2016stacked,li2022simcc,jiang2023rtmpose,purkrabek2025probpose,xu2022vitpose}
first employ a human detector to localize all instances in the input image, followed by single-person pose estimation for each detected instance to predict keypoint locations. 
As a result, their inference time scales approximately linearly with the number of human instances,  
making it inherently challenging to satisfy the requirements of real-time applications.  
Representative works include the HRNet series \cite{sun2019deep,yang2021transpose,yuan2021hrformer},  
SimpleBaseline \cite{xiao2018simple}, and Hourglass \cite{newell2016stacked},  
which typically adopt computationally intensive architectures to achieve high-precision predictions.  
Recent studies have sought to address occlusion and complex scenarios within this paradigm  
\cite{zhang2020distribution,purkrabek2025probpose,sun2024rethinking};  
however, the overall inference pipeline remains constrained by the structural efficiency limitations of the two-stage top-down framework.

Bottom-up methods \cite{cao2019openpose,wei2016convolutional,newell2017associative,kreiss2019pifpaf,
jin2020differentiable,geng2021bottom,cheng2020higherhrnet}
typically predict all keypoints over the entire image in a single forward pass,  
and subsequently employ association algorithms to group them into individual human poses.  
The seminal OpenPose \cite{cao2019openpose} facilitates cross-instance keypoint association by jointly detecting keypoints and modeling Part Affinity Fields (PAFs).  
Recent studies have explored improvements in keypoint representation and supervision strategies to enhance performance  
\cite{luo2021rethinking,qu2023characteristic},  
while others incorporate top-down cues to mitigate the difficulty of keypoint association  
\cite{zhou2023rethinking,zhao2022dpit}.  
Nevertheless, bottom-up approaches generally rely on explicit keypoint grouping and post-processing procedures. This design introduces additional computational overhead during inference, thereby limiting their efficiency in real-time scenarios.

Owing to the inherent processing complexity of two-stage approaches,  
their inference pipelines struggle to satisfy the requirements of real-time pose estimation applications.  
Consequently, single-stage methods capable of directly localizing all human instances and their keypoints from a single image  
have attracted increasing attention  
\cite{nie2019single,xiao2022adaptivepose++,maji2022yolo,lu2024rtmo,mcnally2022rethinking,yang2023explicit}.  
One of the earliest single-stage pose estimation methods,  
Single-Stage Multi-Person Pose Estimation (SPM) \cite{nie2019single},  
estimates keypoint locations via progressive offset prediction,  
pioneering an efficient single-stage paradigm for multi-person pose estimation.  
Subsequent works have integrated Transformer architectures into single-stage pose estimation networks  
\cite{yang2023explicit,shi2022end,tan2024diffusionregpose}.  
However, owing to their complex architectures and substantial computational cost, these methods often fail to meet strict real-time latency constraints.

Inspired by the rapid progress in real-time object detection, particularly the YOLO series  
\cite{redmon2016you,ge2021yolox},  
several single-stage multi-person pose estimation methods have emerged, including YOLO-Pose \cite{maji2022yolo}, CenterPose \cite{duan2019centernet},  
RTMO \cite{lu2024rtmo}, and KAPAO \cite{mcnally2022rethinking}.  
These approaches inherit the design principles and architectural framework of real-time object detectors,  
localizing human instances in a single forward pass  
and predicting the associated keypoints with reference to the instance root position or detection results.  
As the demand for both real-time performance and accuracy in multi-person pose estimation continues to grow,  
single-stage methods have demonstrated competitive advantages over two-stage approaches in terms of inference efficiency and overall performance.

\subsection{Representation Learning for Pose Estimation}
In multi-person pose estimation, the choice of representation for prediction targets plays a critical role in determining model performance.  
An early approach, DeepPose \cite{toshev2014deeppose}, directly regresses keypoint coordinates, formulating pose estimation as an end-to-end coordinate regression problem.  
Subsequent works predominantly model keypoints as pixel-wise probability distributions, i.e., heatmap representations to capture spatial uncertainty.  
This representation has been widely adopted across bottom-up, top-down, and single-stage frameworks  
\cite{newell2016stacked,cao2019openpose,sun2019deep,newell2017associative,
wang2022contextual,li2022simcc,lu2024rtmo,xu2022vitpose,jiang2023rtmpose}.  
Some representative approaches including OpenPose \cite{cao2019openpose}, HRNet \cite{sun2019deep},  
and several single-stage models  
\cite{mao2021fcpose,shi2022end,lu2024rtmo,liu2023group},  
learn Gaussian-based spatial response maps and rely on subsequent decoding procedures to obtain keypoint coordinates.  
Recent studies have further enhanced the representational capacity of heatmaps by increasing output resolution  
\cite{wang2023lightweight}  
or modeling keypoints independently  
\cite{zhou2023rethinking,cao2021dekr}.

Owing to the ability to explicitly model spatial uncertainty,  
heatmap-based representations have generally been regarded as more accurate than direct regression strategies.  
However, such approaches typically rely on downsampled feature maps for prediction,  
which inevitably introduces quantization errors and compromises localization accuracy.  
To mitigate this issue, some methods incorporate learned offset refinement  
\cite{duan2019centernet,gu2021removing}  
or adopt test-time augmentation strategies such as flip-test.  
More recent works, including SimCC \cite{li2022simcc},  
RTMPose \cite{jiang2023rtmpose}, and RTMO \cite{lu2024rtmo},  
further perform probabilistic coordinate modeling in sub-pixel space.  
Although these techniques improve prediction accuracy, they substantially increase representational complexity and computational cost. Moreover, they continue to rely on additional decoding and post-processing steps, which limits their suitability for real-time applications.

In contrast, several studies have explored more direct regression-based formulations for pose estimation.  
For instance, DistilPose \cite{ye2023distilpose} leverages heatmap predictions to supervise regression learning, while RLE \cite{li2021human} incorporates keypoint uncertainty into the regression training process by introducing a normalizing flow model \cite{dinh2016density}, leading to improved performance of regression-based representations.
CenterPose \cite{duan2019centernet} predicts keypoint offset vectors relative to the human center.  
These findings suggest that regression-based representations are more naturally aligned with pose estimation as a structured regression problem.
The empirical accuracy gap observed in earlier regression-based approaches is therefore more plausibly attributed to architectural design and optimization constraints, rather than inherent limitations of regression representations themselves.

Inspired by the success of real-time object detection, recent single-stage approaches adopt multi-scale representations to handle human instances of varying scales within a unified forward pass. Representative methods include YOLO-Pose \cite{maji2022yolo}, RTMO \cite{lu2024rtmo}, and KAPAO \cite{mcnally2022rethinking}, which achieve competitive real-time performance and accuracy.  
From the representation learning perspective, however,  
the regression process in these methods is typically conditioned on detected bounding boxes,  
and keypoint prediction during both training and inference remains governed by box-centric, box-driven modeling assumptions.  
Under such formulation, pose quality is not directly optimized as a primary objective, thereby limiting the representational capacity of regression-based approaches.

Therefore, it is essential to retain the advantages of multi-scale representations while rethinking keypoint representation and organization, and to develop an efficient end-to-end single-stage framework that directly optimizes keypoint regression quality as the primary objective.
\section{Methodology}
\subsection{Architecture Design}
Single-stage pose estimation methods aim to jointly perform human localization and pose coordinate prediction within a unified framework, with the design goals of high inference efficiency, competitive accuracy, and low computational overhead.
In practice, YOLO-like architectures are widely adopted as the backbone of single-stage pose estimation frameworks due to their efficient feature aggregation and multi-scale prediction design.

However, regression-based pose estimation is commonly regarded as difficult to achieve state-of-the-art accuracy.
To improve pose estimation performance, some studies modify keypoint representations, such as transforming regression vectors into probabilistic maps (e.g., RTMO\cite{lu2024rtmo}) or modeling each keypoint as an anchor-based target (e.g., KAPAO\cite{mcnally2022rethinking}).
Other approaches enhance network performance by increasing model capacity and architectural complexity.
However, such strategies often deviate from the original design principles of efficiency and simplicity inherent to single-stage frameworks.

\begin{figure}[!h]
  \centering
  \includegraphics[width=0.48\textwidth]{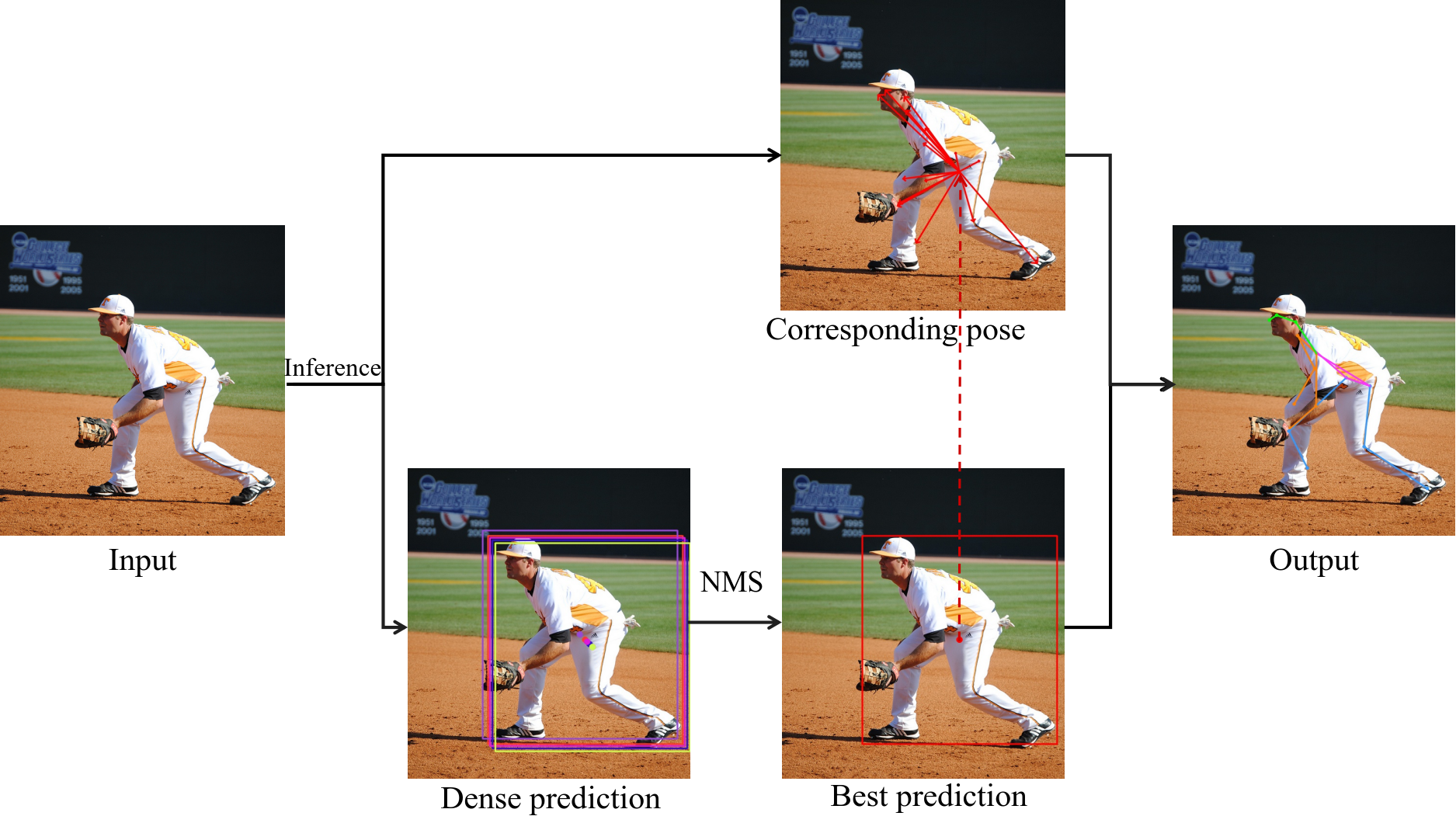}
  \caption{Illustration of the typical box-driven pipeline: the bounding-box-centered single-stage pose estimation paradigm.}
  \label{figure:NMS}
\end{figure}

We re-examine the design of this architecture from a modeling perspective. Within existing frameworks, positive sample selection for pose estimation is tightly coupled with the parallel human detection branch, such that pose regression is conditioned on bounding box localization results. In practice, positive samples are determined by detection confidence scores followed by NMS filtering, after which keypoint vectors are regressed at the selected locations.
The baseline inference process is illustrated in \cref{figure:NMS}.

This IoU-based target selection mechanism introduces structural limitations by coupling pose regression with bounding-box localization. Moreover, the backbone and neck are typically designed to extract generic features for multiple tasks, whereas pose estimation requires richer high-level semantic representations than bounding-box detection. Joint supervision of box detection and pose estimation within a single network may compromise pose accuracy, as the backbone cannot fully specialize in features dedicated to pose semantics.

Based on this analysis, we propose a regression-based pose estimation framework that removes the box prediction branch and adopts a fully keypoint-driven design.

In dense prediction settings, instance selection is typically performed via NMS. However, owing to the increased structural complexity of pose estimation, applying NMS to pose predictions introduces additional computational overhead, thereby reducing inference efficiency \cite{zhao2024detrs}.

To enable direct and efficient selection of high-quality pose predictions, we draw inspiration from the training strategy of YOLO-v10 \cite{wang2024yolov10} and incorporate a dual-head assignment mechanism into pose estimation.
Specifically, a Multi-Assignment Head (MAH) is employed to enable dense supervision during training. For each human instance, MAH assigns multiple positive samples, providing richer and more stable training signals.
During inference, only the Single-Assignment Head (SAH) is retained. During training, SAH assigns a single positive sample per instance and is jointly optimized with MAH.
This design enables SAH to directly localize sparse and high-quality grid locations at inference time, thereby facilitating reliable pose regression without NMS.
Building upon this design, we propose a new Keypoint-Centric network paradigm characterized as follows:
\begin{itemize}
  \item Following confidence-guided positive sample selection, keypoint offset vectors are directly regressed from the pose branch, enabling feature learning to specialize exclusively in pose semantics. The prediction head is redesigned to better accommodate the representational demands of pose estimation.
  \item The dual-head assignment strategy is adapted to pose estimation to directly localize high-quality pose grid locations, thereby enabling NMS-free inference without additional post-processing.
\end{itemize}

\begin{figure*}[!t]
  \centering
  \includegraphics[width=0.95\textwidth]{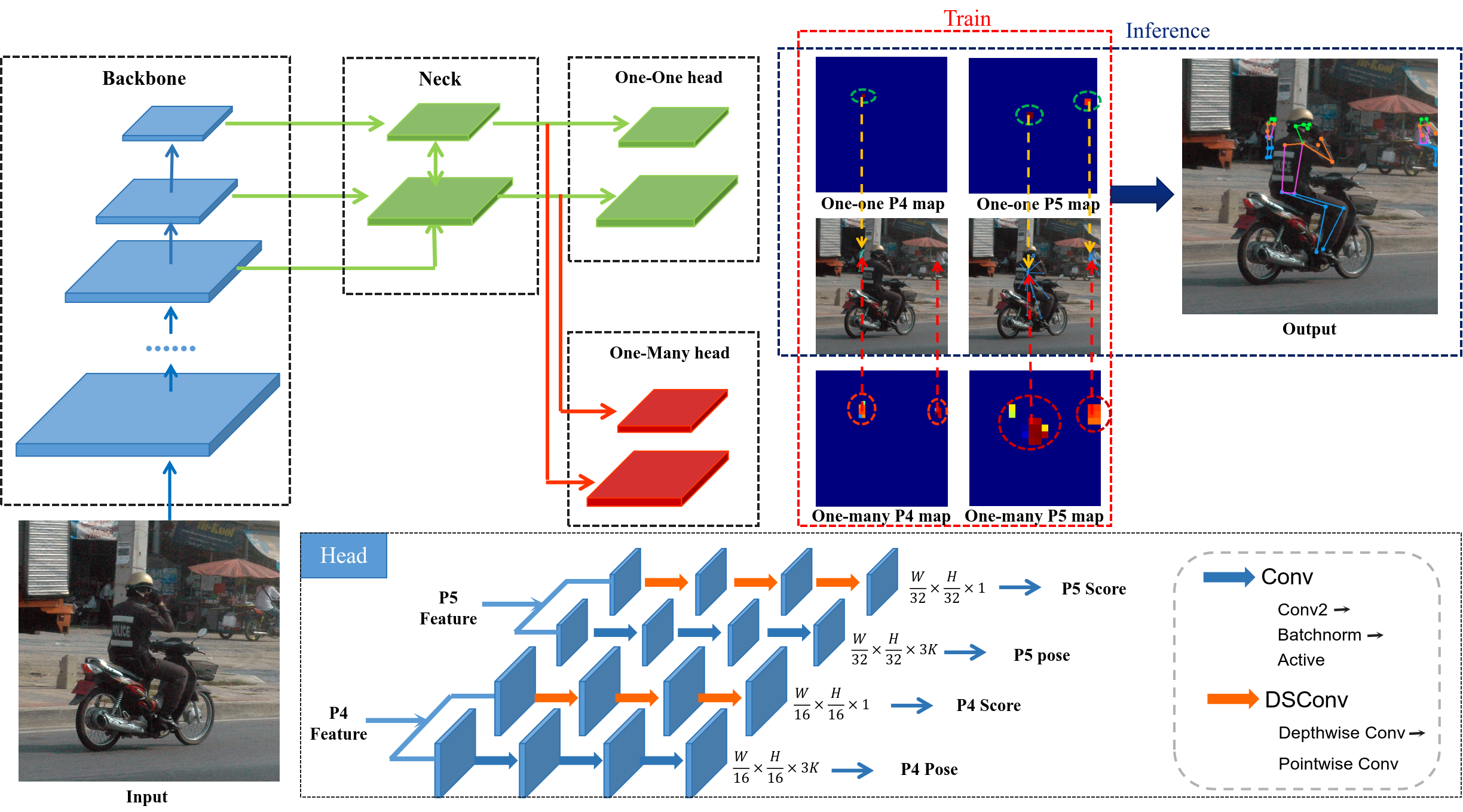}
  \caption{Overview of the ER-Pose framework. ER-Pose adopts keypoint-driven single-stage architecture. Feature extraction design follows YOLOv8, with CSPDarkNet\cite{wang2020cspnet,bochkovskiy2020yolov4} used as backbone and PAN\cite{liu2018path} employed as neck to produce multi-scale features. Head consists of human confidence branch and pose branch. Confidence branch adopts DSConv\cite{howard2017mobilenets} to reduce computational cost, while keypoint regression employs standard convolution to preserve sufficient representational capacity. Network is trained under MAH-SAH scheme.}
  \label{figure:framework}
\end{figure*}

During training, we introduce a keypoint-oriented assignment metric to refine positive sample selection, as detailed in \cref{sec:misalignment}.
We further propose a smoother keypoint regression loss, termed Smooth-OKS, to enhance global supervision, as described in \cref{sec:oks_loss}.
The resulting model, termed Efficient Regression Pose (ER-Pose), forms a fully regression-based end-to-end inference framework, as illustrated in \cref{figure:framework}.

\subsection{Task Misalignment and Keypoint-driven Assignment Metric}
\label{sec:misalignment}
To improve inference efficiency, networks typically rely on backbone features shared across multiple parallel task branches. Within box-driven single-stage pose estimation frameworks, the confidence branch, bounding-box branch, and pose estimation branch operate in parallel and share multi-scale features fused by the backbone and neck modules.

\begin{figure}[!h]
  \centering
  \includegraphics[width=0.45\textwidth]{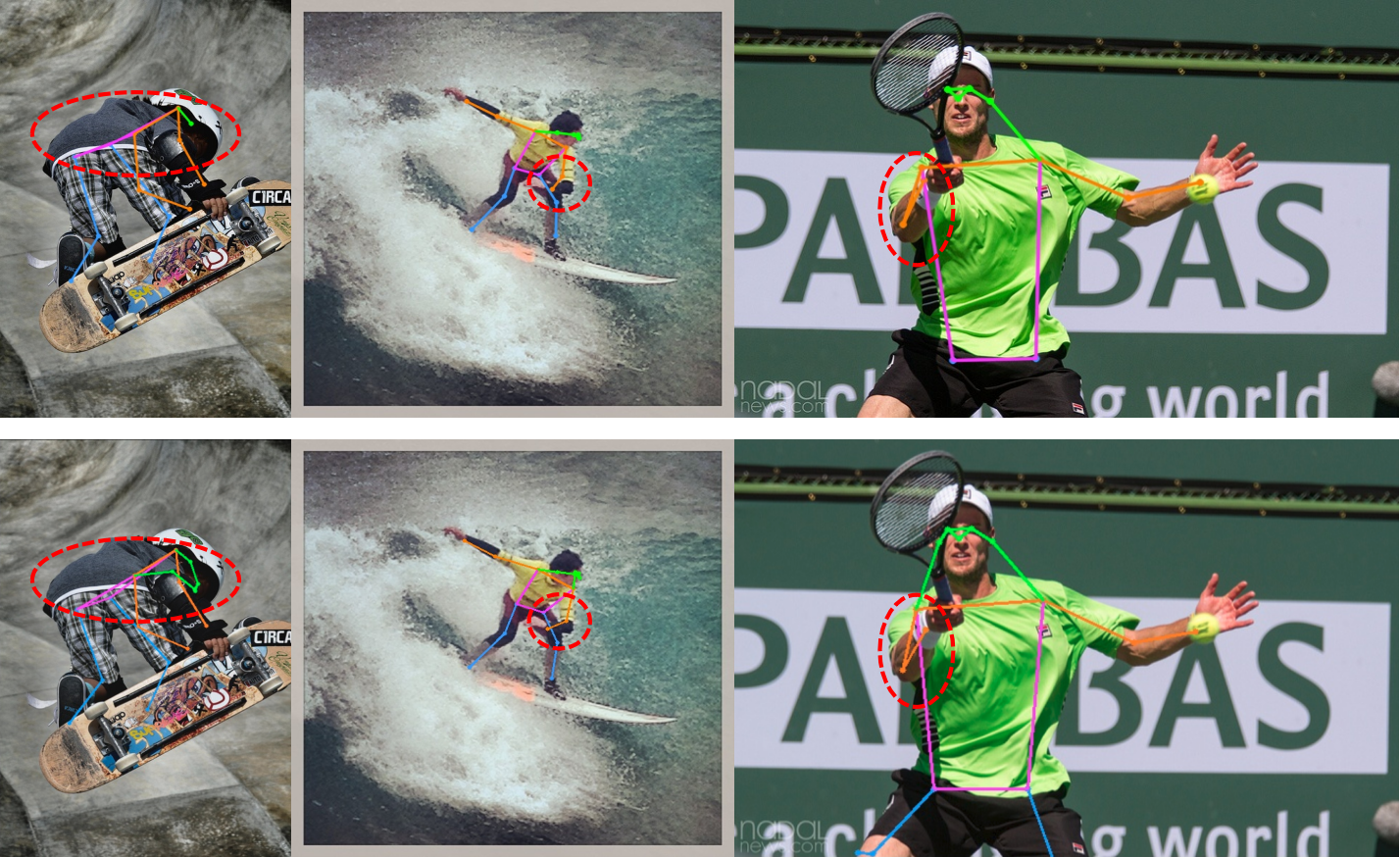}
  \caption{Suboptimal keypoint feature selection induced by box-driven modeling. The upper part shows positive samples selected by the network, while the lower part presents samples from the same network that yield more accurate keypoint estimates but are suppressed during inference.}
  \label{figure:Box vs pose}
\end{figure}

Due to the multi-scale feature design, these frameworks commonly adopt dynamic target assignment strategies during training to determine positive sample locations. In existing YOLO-like baselines, positive sample selection is primarily guided by bounding-box IoU, with locations exhibiting larger IoU values favored during both training and inference\cite{ultralytics_yolov8_pose,feng2021tood}. As keypoint feature selection is largely governed by IoU-based criteria, the pose branch may receive feature representations that are not fully aligned with keypoint regression, leading to a discrepancy between extracted pose features and the regression objective. This phenomenon is illustrated in \cref{figure:Box vs pose}.

To improve pose estimation accuracy, pose regression should be decoupled from bounding-box localization, and dynamic assignment must be reformulated with pose accuracy as the primary optimization objective. Accordingly, after removing the bounding-box branch from the network, the keypoint assignment metric used during training is defined as follows:
\begin{equation}
  Score=conf^{\alpha}\times OKS^{\beta}
\end{equation}
where $conf$ denotes the confidence score of the human instance, and 
$OKS$ corresponds to the OKS score computed from the pose branch, defined as:
\begin{equation}
  \label{fun:OKS}
  \mathrm{OKS}=\frac{\sum_{i}exp{({-d_i^2/2s^2k_i^2})}\delta(v_i>0)}{\sum_{i}\delta(v_i>0)}
\end{equation}
where $d_i$ denotes the Euclidean distance between the predicted and ground-truth locations of the $i$-th keypoint; $s$ denotes the scale factor of the target human, defined as the square root of the area of the human bounding box; $k_i$ represents the importance weight associated with each keypoint, characterizing variations in keypoint importance and scale uncertainty; $v_i$ denotes the visibility annotation of the $i$-th keypoint, where $v_i=0$ indicates invisibility, $v_i=1$ indicates visibility but low reliability, and $v_i=2$ indicates visibility with high reliability. $\delta(v_i)$ serves to filter invisible keypoints and is defined as follows:
\begin{equation}
    \delta(v_i)=\begin{cases}
        1, & v_i>0 \\
        0, & v_i=0
    \end{cases}
\end{equation}
During training, the $k$ locations with the highest $Score$ values are selected as positive samples for training supervision. The algorithm proceeds as follows:
\begin{algorithm}[!h]
  \caption{keypoint-driven Task Assignment in ER-Pose (per-GT Top-$K$)}
  \label{alg:erpose_assignment_per_gt}
  \begin{algorithmic}[1]
  \REQUIRE Input image $I$, ground-truth humans $G$
  \ENSURE Positive grid set $S$
  \STATE $S \leftarrow \emptyset$
  \STATE Forward $I$ through ER-Pose to obtain grid predictions $P$
  \FOR{each ground-truth target $g \in G$}
  \FOR{each grid prediction $p \in P$}
          \STATE Obtain confidence score $conf_p$
          \STATE Compute $\mathrm{OKS}(p, g)$
          \STATE $Score_{p,g} \leftarrow conf_p^{\alpha} \cdot \mathrm{OKS}(p, g)^{\beta}$
      \ENDFOR
      \STATE Select top $K$ grids $\mathcal{P}_g \subseteq P$ by $Score_{p,g}$ (descending)
      \STATE $S \leftarrow S \cup \mathcal{P}_g$
  \ENDFOR
  \RETURN $S$
  \end{algorithmic}
\end{algorithm}

\begin{figure}[!h]
  \centering
  \includegraphics[width=0.48\textwidth]{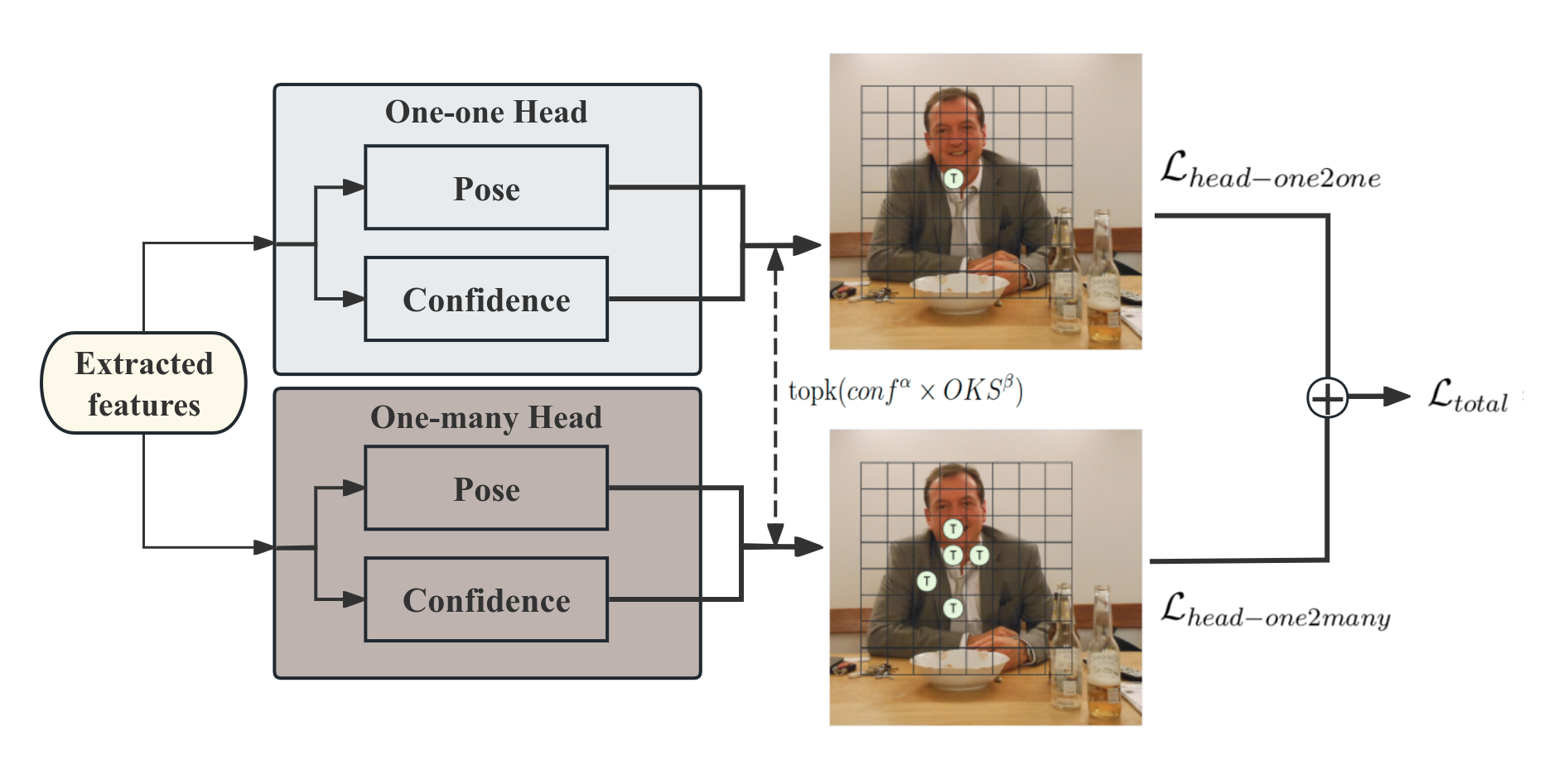}
  \caption{The assignment paradigm}
  \label{figure:MAH-SAH}
\end{figure}

In conjunction with the dual-head architecture, the assignment paradigm is illustrated in \cref{figure:MAH-SAH}.

The TOOD\cite{feng2021tood} framework first identified the issue of sample assignment misalignment. However, the discussion remains primarily conceptual and lacks a formal quantitative analysis.
To quantitatively analyze consistency between sample assignment results and pose estimation objectives, Task Alignment Error is formulated mathematically in this work. Let an image contain $N$ pose annotations denoted by $\mathcal{G}={g_i}{i=1}^{N}$, and let the network produce $M$ pose candidate predictions denoted by $\hat{\mathcal{Y}}={\hat y_j}{j=1}^{M}$. Define similarity matrix $S\in\mathbb{R}^{N\times M}$, where $S_{ij}\triangleq \mathrm{OKS}(g_i,\hat y_j)$.
The globally optimal one-to-one matching under the OKS criterion is solved over the set of injective mappings from ${1,\dots,N}$ to ${1,\dots,M}$, denoted by $\mathcal{I}(N,M)$:
\begin{equation}
    \begin{aligned}
      P^\star 
      &\in \arg\max_{P\in \mathcal{I}(N,M)}
      \sum_{i=1}^{N} \mathrm{OKS}\!\left(g_i,\hat y_{P(i)}\right),\\
      \mathcal{I}(N,M)
      &\triangleq
      \{P:\{1,\dots,N\}\to\{1,\dots,M\}\mid P \text{ is injective}\}.
    \end{aligned}
\end{equation}

Accordingly, the OKS-optimal prediction corresponding to $g_i$ is defined as $\hat y_i^\star \triangleq \hat y_{P^\star(i)}$.
On the other hand, the positive sample prediction actually selected during inference is denoted as $\hat y_i^{s}$.
Task Alignment Error is defined as:
\begin{equation}
\label{fun:TAE}
\mathrm{TAE}
\triangleq
\frac{1}{N}\sum_{i=1}^{N}
\Big(
\mathrm{OKS}(g_i,\hat y_i^\star)-\mathrm{OKS}(g_i,\hat y_i^{s})
\Big).
\end{equation}
This metric characterizes the extent to which the sample assignment strategy deviates from the pose-optimal solution in OKS space.
When $\mathrm{TAE}=0$, the positive samples selected during inference exactly match the OKS-optimal assignment.
Larger $\mathrm{TAE}$ values indicate greater deviation from the pose-optimal solution, thereby introducing stronger suboptimality in pose regression.
This error does not participate in training optimization and serves solely as an analytical metric, with statistical results reported in \cref{sec: assignment error}.

\subsection{Loss Function}
\label{sec:oks_loss}

To better exploit the modeling capacity of a keypoint-driven single-stage regression framework, supervision should be applied directly to structured pose representations rather than to independent coordinate components or auxiliary detection-related variables.
In existing regression-based pose estimation approaches, keypoint supervision commonly relies on coordinate-error-based losses, such as L1, L2, and Smooth L1 losses. 
These loss functions primarily measure point-wise geometric discrepancies and do not explicitly account for variations in keypoint scale or regression difficulty. Consequently, they offer limited capacity to model human pose as a high-dimensional structured prediction problem.

In this work, following the standard Object Keypoint Similarity (OKS) metric defined in \cref{fun:OKS},
the normalized error for the $i$-th keypoint is defined as 
\begin{equation}
    u_i=\frac{d_i}{sk_i}
\end{equation}
where $d_i$ denotes the Euclidean distance between the predicted and ground-truth locations of the $i$-th keypoint, $s$ represents the scale of the corresponding human instance, and $k_i$ is a keypoint-specific normalization constant.
The term $k_i$ reflects variations in spatial stability and annotation noise across different keypoints and is typically predefined for each dataset. In the absence of dataset-specific statistics, $k_i=\frac{1}{K}$ is adopted.

\begin{figure}[!h]
  \centering
  \begin{subfigure}{0.14\textwidth}
      \centering
      \includegraphics[width=\linewidth]{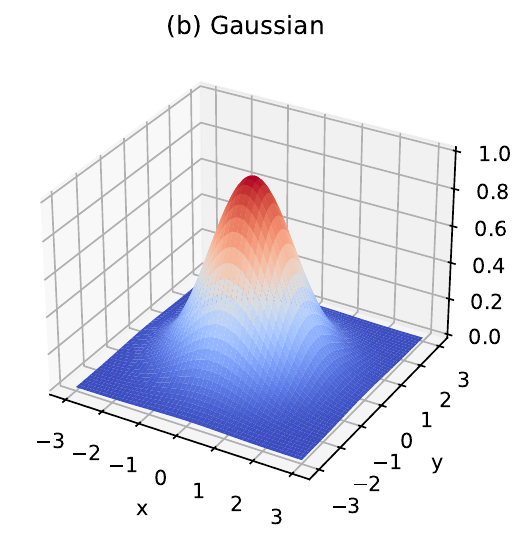}
      \caption{Gaussian}
      \label{figure:Gaussian}
  \end{subfigure}
  \begin{subfigure}{0.14\textwidth}
      \centering
      \includegraphics[width=\linewidth]{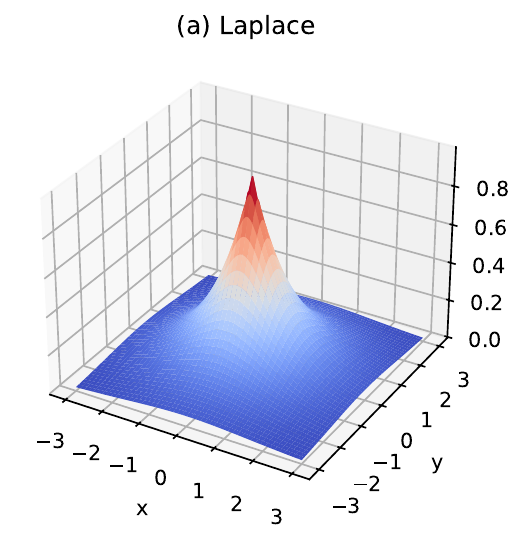}
      \caption{Laplace}
      \label{figure:Laplace}
  \end{subfigure}
  \begin{subfigure}{0.14\textwidth}
    \centering
    \includegraphics[width=\linewidth]{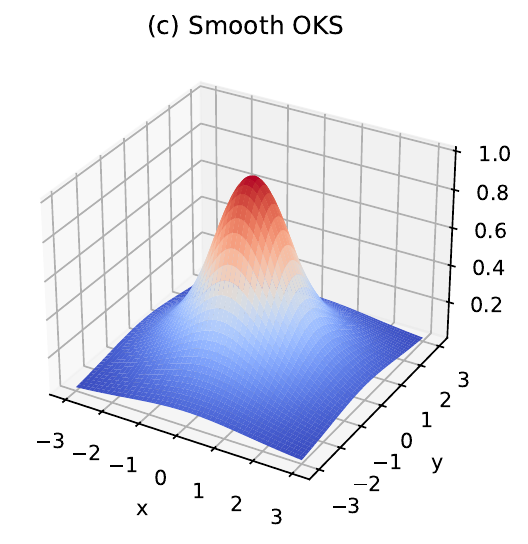}
    \caption{SOKS}
    \label{figure:Smooth-OKS}
  \end{subfigure}
  \caption{Comparison of Gaussian, Laplace, and SOKS distributions.}
  \label{figure:distribution}
\end{figure}

During training, keypoint predictions are optimized under an OKS-consistent objective, formulated as:
\begin{equation}
  \mathcal{L}_{oks}=1-\frac{\sum_{i} \exp{({-u_i^2/2})}\delta(v_i>0)}{\sum_{i}\delta(v_i>0)}
\end{equation}
This formulation corresponds to imposing a Gaussian constraint in the normalized error space. In regression-based pose estimation, large-error samples are common during early training stages. Under Gaussian constraints, supervisory signals decay rapidly for such samples, leading to vanishing gradient contributions. To improve optimization stability in the large-error regime while preserving fine-grained supervision in the low-error regime, a Laplace-type constraint with stronger gradient responses is applied to large errors, whereas the Gaussian form is retained for small errors to ensure stable refinement of precise pose regression. A smooth transition between the two distributions ensures balanced error modeling and stable optimization dynamics.
The resulting formulation is defined as:
\begin{equation}
  \begin{cases}
      -e^{\frac{u^2}{2} }, & u^2 < 1 \\
      -e^{\frac{2\sqrt{u^2}-1}{2} }, & u^2 \geq 1
  \end{cases}
\end{equation}
This formulation is termed Smooth-OKS (SOKS). \cref{figure:distribution} illustrates the response curves of Gaussian, Laplace, and SOKS constraints in the normalized error space, highlighting their differences in prediction similarity behavior.

Keypoint predictions are supervised under the SOKS distribution, and the corresponding loss is defined as:
\begin{equation}
  \mathcal{L}_{Pose}=1-SOKS
\end{equation}

Human confidence prediction and keypoint visibility prediction are optimized using binary cross-entropy loss with logits, defined as:
\begin{equation}
  \mathcal{L}_{\mathrm{vis}} = - \left( v_i \log(\hat{v}_i) + (1 - v_i) \log(1 - \hat{v}_i) \right)
\end{equation}
\begin{equation}
  \mathcal{L}_{\mathrm{conf}} = - \left( c \log(\hat{p}_{\mathrm{conf}}) + (1 - c) \log(1 - \hat{p}_{\mathrm{conf}}) \right)
\end{equation}
where $v_i \in {0,1}$ denotes the ground-truth visibility of the $i$-th keypoint, with 1 indicating visibility and 0 indicating invisibility. The predicted visibility probability is denoted by $\hat{v}_i \in (0,1)$.
The variable $c \in {0,1}$ represents the ground-truth class label of the instance, where 1 corresponds to a human target and 0 corresponds to background. The predicted human confidence score at the corresponding grid location is denoted by $\hat{p}_{\mathrm{conf}} \in (0,1)$.
Finally, the three losses are linearly combined with weighting coefficients to form the overall loss function, defined as:
\begin{equation}
    \mathcal{L}_{\mathrm{grid}} = 
    \lambda_{\mathrm{pose}} \mathcal{L}_{\mathrm{pose}} +
    \lambda_{\mathrm{vis}} \mathcal{L}_{\mathrm{vis}} +
    \lambda_{\mathrm{conf}} \mathcal{L}_{\mathrm{conf}}
\end{equation}
$\lambda_{\mathrm{pose}}$, $\lambda_{\mathrm{vis}}$, and $\lambda_{\mathrm{conf}}$ denote the weighting coefficients for keypoint loss, visibility loss, and human confidence loss, respectively.
In our implementation, they are set to $15:1:1$.

\section{Experiments}
\subsection{Experimental Settings}
\noindent\textbf{Datasets.} Experiments were primarily conducted on the COCO\cite{lin2014microsoft} dataset, which contains over 200,000 images and approximately 250,000 human instances annotated with 17 keypoints. The train2017 split comprises more than 50,000 images, while val2017 and test-dev contain 5,000 and 20,000 images, respectively. Models were trained on train2017 and evaluated on val2017 and test-dev. Additional evaluations were conducted on the CrowdPose\cite{li2019crowdpose} and OCHuman\cite{zhang2019pose2seg} datasets to assess generalization performance.

\noindent\textbf{Evaluation metrics.}
Reported results include $mAP$, $mAR$, threshold-specific metrics (e.g., $AP_{50}$, $AP_{75}$), scale-specific metrics (e.g., $AP_S$, $AP_M$, $AP_L$), as well as difficulty-level metrics (e.g., $AP_E$, $AP_M$, and $AP_H$ on CrowdPose), following the standard evaluation protocols. In addition, computational resource metrics, such as parameter count and inference speed, are reported for comprehensive comparison.

\noindent\textbf{Training details.}
All experiments were conducted on Ubuntu 20.04, with training performed on 4 NVIDIA RTX 3090 GPUs. Data augmentation includes random color jittering, random geometric transformations, and Mosaic augmentation \cite{bochkovskiy2020yolov4}. Unless otherwise stated, input images are resized to $640 \times 640$. The initial learning rate is set to 0.01 and gradually decayed to 0.0001 using cosine annealing \cite{loshchilov2016sgdr}. Optimization is performed using AdamW \cite{loshchilov2017decoupled} with the weight decay of 0.05 for 600 epochs. All models are implemented in PyTorch.

\noindent\textbf{Inference implementation.}
For fair comparison, inference latency was measured on an NVIDIA Tesla V100 GPU using ONNX Runtime under FP32 precision with batch size set to 1.

\subsection{Comparison with State-of-the-Art Methods on COCO}
To assess the effectiveness of the proposed keypoint-driven design framework, network performance was evaluated on the COCO test-dev dataset, with parameter count, accuracy, recall, and inference latency reported. Comparisons were conducted with representative state-of-the-art pose estimation methods, including HigherHRNet \cite{cheng2020higherhrnet}, DEKR \cite{cao2021dekr}, KAPAO \cite{mcnally2022rethinking}, RTMO \cite{lu2024rtmo}, CID \cite{wang2022contextual}, CenterNet \cite{duan2019centernet}, and ED-Pose \cite{yang2023explicit}. Results are summarized in \cref{table:compare}.
\begin{table*}[!t]
  \centering
  \renewcommand{\arraystretch}{1.1}
  \setlength{\tabcolsep}{12pt}
  \caption{Comparison results of ER-Pose with advanced methods on the COCO test-dev set.
Inference latency was measured on one Tesla V100 GPU under identical hardware conditions,
with evaluation performed using onnxruntime.}
  \begin{tabular}{cccccccc}
  \hline
  \multicolumn{1}{c|}{Method} & \multicolumn{1}{c|}{Backbone} & \multicolumn{1}{c|}{Input Size} & \multicolumn{1}{c|}{Params (M)} & \multicolumn{1}{c|}{Latency (ms)} & \multicolumn{1}{c|}{AP} & \multicolumn{1}{c|}{$AP_{50}$} & \multicolumn{1}{c}{$AP_{75}$} \\ \hline
  \multicolumn{1}{c|}{HigherHRNet\cite{cheng2020higherhrnet}}       & \multicolumn{1}{c|}{HRNet-W32\cite{sun2019deep}}         & \multicolumn{1}{c|}{512$\times$512}          & \multicolumn{1}{c|}{28.6}       & \multicolumn{1}{c|}{34.398}     & \multicolumn{1}{c|}{66.4}   & \multicolumn{1}{c|}{87.5}     & \multicolumn{1}{c}{72.8}             \\ 
  \multicolumn{1}{c|}{CenterNet\cite{duan2019centernet}}       & \multicolumn{1}{c|}{Hourglass-3x\cite{newell2016stacked}}         & \multicolumn{1}{c|}{512$\times$512}          & \multicolumn{1}{c|}{194.9}       & \multicolumn{1}{c|}{45.855}     & \multicolumn{1}{c|}{63.0}   & \multicolumn{1}{c|}{86.8}     & \multicolumn{1}{c}{69.6}             \\ 
  \multicolumn{1}{c|}{DEKR\cite{cao2021dekr}}       & \multicolumn{1}{c|}{HRNet-W32\cite{sun2019deep}}         & \multicolumn{1}{c|}{512$\times$512}          & \multicolumn{1}{c|}{29.6}       & \multicolumn{1}{c|}{33.935}     & \multicolumn{1}{c|}{68.0}   & \multicolumn{1}{c|}{86.7}     & \multicolumn{1}{c}{74.5} \\           
  \multicolumn{1}{c|}{DEKR\cite{cao2021dekr}}       & \multicolumn{1}{c|}{HRNet-W48\cite{sun2019deep}}         & \multicolumn{1}{c|}{640$\times$512}          & \multicolumn{1}{c|}{65.7}       & \multicolumn{1}{c|}{36.585}     & \multicolumn{1}{c|}{71.0}   & \multicolumn{1}{c|}{88.3}     & \multicolumn{1}{c}{77.4} \\\hline
  \multicolumn{1}{c|}{CID\cite{wang2022contextual}}       & \multicolumn{1}{c|}{HRNet-W32\cite{sun2019deep}}         & \multicolumn{1}{c|}{512$\times$512}          & \multicolumn{1}{c|}{29.4}       & \multicolumn{1}{c|}{45.678}     & \multicolumn{1}{c|}{68.9}   & \multicolumn{1}{c|}{88.5}     & \multicolumn{1}{c}{76.6}        \\\hline
  \multicolumn{1}{c|}{ED-Pose\cite{yang2023explicit}}       & \multicolumn{1}{c|}{ResNet-50\cite{he2016deep}}         & \multicolumn{1}{c|}{800 $\times$ 1333}          & \multicolumn{1}{c|}{50.6}       & \multicolumn{1}{c|}{171.971}     & \multicolumn{1}{c|}{69.8}   & \multicolumn{1}{c|}{90.2}     & \multicolumn{1}{c}{77.2}            \\\hline
  \multicolumn{1}{c|}{KAPAO-s\cite{mcnally2022rethinking}}       & \multicolumn{1}{c|}{CSPNet\cite{wang2020cspnet}}         & \multicolumn{1}{c|}{1280$\times$1280}          & \multicolumn{1}{c|}{12.6}       & \multicolumn{1}{c|}{16.69}     & \multicolumn{1}{c|}{63.8}   & \multicolumn{1}{c|}{88.4}     & \multicolumn{1}{c}{70.4}             \\
  \multicolumn{1}{c|}{KAPAO-m\cite{mcnally2022rethinking}}       & \multicolumn{1}{c|}{CSPNet\cite{wang2020cspnet}}         & \multicolumn{1}{c|}{1280$\times$1280}          & \multicolumn{1}{c|}{35.8}       & \multicolumn{1}{c|}{20.186}     & \multicolumn{1}{c|}{68.8}   & \multicolumn{1}{c|}{90.5}     & \multicolumn{1}{c}{76.5}             \\
  \multicolumn{1}{c|}{KAPAO-l\cite{mcnally2022rethinking}}       & \multicolumn{1}{c|}{CSPNet\cite{wang2020cspnet}}         & \multicolumn{1}{c|}{1280$\times$1280}          & \multicolumn{1}{c|}{77.0}       & \multicolumn{1}{c|}{26.343}    & \multicolumn{1}{c|}{70.3}   & \multicolumn{1}{c|}{91.2}     & \multicolumn{1}{c}{77.8}             \\ \hline
  \multicolumn{1}{c|}{RTMO-s\cite{lu2024rtmo}}       & \multicolumn{1}{c|}{CSPDarknet\cite{bochkovskiy2020yolov4}}         & \multicolumn{1}{c|}{[480,800]}          & \multicolumn{1}{c|}{9.9}       & \multicolumn{1}{c|}{16.827}     & \multicolumn{1}{c|}{66.9}   & \multicolumn{1}{c|}{88.8}     & \multicolumn{1}{c}{73.6}             \\
  \multicolumn{1}{c|}{RTMO-m\cite{lu2024rtmo}}       & \multicolumn{1}{c|}{CSPDarknet\cite{bochkovskiy2020yolov4}}         & \multicolumn{1}{c|}{[480,800]}          & \multicolumn{1}{c|}{22.6}       & \multicolumn{1}{c|}{18.286}     & \multicolumn{1}{c|}{70.1}   & \multicolumn{1}{c|}{90.6}     & \multicolumn{1}{c}{77.1}             \\
  \multicolumn{1}{c|}{RTMO-l\cite{lu2024rtmo}}       & \multicolumn{1}{c|}{CSPDarknet\cite{bochkovskiy2020yolov4}}         & \multicolumn{1}{c|}{[480,800]}          & \multicolumn{1}{c|}{44.8}       & \multicolumn{1}{c|}{19.977}    & \multicolumn{1}{c|}{71.6}   & \multicolumn{1}{c|}{91.1}     & \multicolumn{1}{c}{79.0}             \\ \hline
  \multicolumn{1}{c|}{YOLOv8-Pose-n\cite{ultralytics_yolov8_pose}}       & \multicolumn{1}{c|}{CSPDarknet\cite{bochkovskiy2020yolov4}}         & \multicolumn{1}{c|}{640$\times$640}          & \multicolumn{1}{c|}{3.3}   & \multicolumn{1}{c|}{16.054}     & \multicolumn{1}{c|}{49.1}     & \multicolumn{1}{c|}{80.0}         & \multicolumn{1}{c}{52.5}        \\
  \multicolumn{1}{c|}{YOLOv8-Pose-s\cite{ultralytics_yolov8_pose}}       & \multicolumn{1}{c|}{CSPDarknet\cite{bochkovskiy2020yolov4}}         & \multicolumn{1}{c|}{640$\times$640}          & \multicolumn{1}{c|}{11.6}  & \multicolumn{1}{c|}{17.125}      & \multicolumn{1}{c|}{58.1}     & \multicolumn{1}{c|}{85.7}        & \multicolumn{1}{c}{63.8}        \\
  \multicolumn{1}{c|}{YOLOv8-Pose-m\cite{ultralytics_yolov8_pose}}       & \multicolumn{1}{c|}{CSPDarknet\cite{bochkovskiy2020yolov4}}         & \multicolumn{1}{c|}{640$\times$640}          & \multicolumn{1}{c|}{26.4}  & \multicolumn{1}{c|}{17.491}      & \multicolumn{1}{c|}{63.0}     & \multicolumn{1}{c|}{88.2}       & \multicolumn{1}{c}{70.1}         \\
  \multicolumn{1}{c|}{YOLOv8-Pose-l\cite{ultralytics_yolov8_pose}}       & \multicolumn{1}{c|}{CSPDarknet\cite{bochkovskiy2020yolov4}}         & \multicolumn{1}{c|}{640$\times$640}          & \multicolumn{1}{c|}{44.4}  & \multicolumn{1}{c|}{21.378}      & \multicolumn{1}{c|}{65.5}     & \multicolumn{1}{c|}{89.2}      & \multicolumn{1}{c}{73.4}        \\ \hline
  \multicolumn{1}{c|}{ER-Pose-n} & \multicolumn{1}{c|}{CSPDarknet\cite{bochkovskiy2020yolov4}}         & \multicolumn{1}{c|}{640$\times$640}          & \multicolumn{1}{c|}{\textbf{2.53}}       & \multicolumn{1}{c|}{\textbf{10.587}}     & \multicolumn{1}{c|}{56.0}   & \multicolumn{1}{c|}{84.7}     & \multicolumn{1}{c}{60.5}    \\
  \multicolumn{1}{c|}{ER-Pose-s} & \multicolumn{1}{c|}{CSPDarknet\cite{bochkovskiy2020yolov4}}         & \multicolumn{1}{c|}{640$\times$640}          & \multicolumn{1}{c|}{9.69}       & \multicolumn{1}{c|}{12.544}     & \multicolumn{1}{c|}{64.2}   & \multicolumn{1}{c|}{88.8}         & \multicolumn{1}{c}{71.1}         \\
  \multicolumn{1}{c|}{ER-Pose-m} & \multicolumn{1}{c|}{CSPDarknet\cite{bochkovskiy2020yolov4}}         & \multicolumn{1}{c|}{640$\times$640}          & \multicolumn{1}{c|}{23.36}       & \multicolumn{1}{c|}{14.916}     & \multicolumn{1}{c|}{68.2}   & \multicolumn{1}{c|}{90.7}          & \multicolumn{1}{c}{75.9}      \\
  \multicolumn{1}{c|}{ER-Pose-l} & \multicolumn{1}{c|}{CSPDarknet\cite{bochkovskiy2020yolov4}}         & \multicolumn{1}{c|}{640$\times$640}          & \multicolumn{1}{c|}{39.95}       & \multicolumn{1}{c|}{16.948}     & \multicolumn{1}{c|}{70.2}   & \multicolumn{1}{c|}{91.3}          & \multicolumn{1}{c}{77.6}     \\ \hline
  \multicolumn{1}{c|}{ER-Pose-n} & \multicolumn{1}{c|}{CSPDarknet\cite{bochkovskiy2020yolov4}}         & \multicolumn{1}{c|}{800$\times$800}          & \multicolumn{1}{c|}{\textbf{2.53}}       & \multicolumn{1}{c|}{11.844}     & \multicolumn{1}{c|}{58.1}   & \multicolumn{1}{c|}{86.3}     & \multicolumn{1}{c}{63.4}    \\
  \multicolumn{1}{c|}{ER-Pose-s} & \multicolumn{1}{c|}{CSPDarknet\cite{bochkovskiy2020yolov4}}         & \multicolumn{1}{c|}{800$\times$800}          & \multicolumn{1}{c|}{9.69}       & \multicolumn{1}{c|}{14.455}     & \multicolumn{1}{c|}{66.6}   & \multicolumn{1}{c|}{90.0}         & \multicolumn{1}{c}{73.6}         \\
  \multicolumn{1}{c|}{ER-Pose-m} & \multicolumn{1}{c|}{CSPDarknet\cite{bochkovskiy2020yolov4}}         & \multicolumn{1}{c|}{800$\times$800}          & \multicolumn{1}{c|}{23.36}       & \multicolumn{1}{c|}{16.255}     & \multicolumn{1}{c|}{70.3}   & \multicolumn{1}{c|}{91.4}          & \multicolumn{1}{c}{77.9}      \\
  \multicolumn{1}{c|}{ER-Pose-l} & \multicolumn{1}{c|}{CSPDarknet\cite{bochkovskiy2020yolov4}}         & \multicolumn{1}{c|}{800$\times$800}          & \multicolumn{1}{c|}{39.95}       & \multicolumn{1}{c|}{19.411}     & \multicolumn{1}{c|}{\textbf{71.9}}   & \multicolumn{1}{c|}{\textbf{91.9}}          & \multicolumn{1}{c}{\textbf{79.6}}     \\ \hline
  \end{tabular}
  \label{table:compare}
\end{table*}

Under comparable parameter budgets, the four model scales of ER-Pose consistently achieve faster inference speeds than existing state-of-the-art methods.
This efficiency stems from the fully regression-based inference paradigm, which eliminates the need for any post-processing operations.
Moreover, ER-Pose attains higher accuracy at lower inference latency, demonstrating a favorable efficiency–accuracy trade-off.

At an input resolution of 640,
ER-Pose outperforms its baseline YOLO-Pose by +6.9, +6.1, +5.2, and +4.7 mAP across the four model scales respectively.
When the input resolution is increased to 800,
ER-Pose-l achieves 71.9 AP with an inference latency of 19.41 ms,
surpassing other state-of-the-art models in accuracy while maintaining competitive real-time performance.

Furthermore,
we increased the input resolution to 960 to evaluate the effect of higher input resolution on model performance,
with results reported on the COCO val2017 subset.
All experiments were conducted using ER-Pose-n, and the corresponding results are summarized in \cref{tab:input size experiments}.
\begin{table}[!h]
  \centering
  \renewcommand{\arraystretch}{1.4}
  \setlength{\tabcolsep}{10pt}
  \caption{Comparison of model performance under different input Resolutions}
  \label{tab:input size experiments}
  \begin{tabular}{cccccc}
  \hline
  Input size & Latency(ms) & AP & $\mathrm{AP}_{50}$ & $\mathrm{AP}_{75}$ & AR \\ \hline
  640        & 10.587  & 57.1 & 84.6  &62.0  &62.4 \\
  800        & 11.844  & 59.5 &86.2   &65.0  &64.5    \\ 
  960        & 13.412  & 60.9 &86.8   &68.8  &65.9    \\ \hline
  \end{tabular}
\end{table}
Increasing the input resolution yields consistent gains in AP. Specifically, scaling the resolution from 640 to 800 and further to 960 improves AP by +2.4 and +1.4 points respectively. However, these gains come at the cost of increased inference latency, thereby reducing real-time efficiency.

\subsection{Structural Ablation Study}
The ER-Pose architecture adopts CSPDarkNet \cite{wang2020cspnet} as the backbone and Path Aggregation Network (PAN) \cite{liu2018path} as the neck, leveraging multi-scale feature maps to model human instances of varying scales. PAN generates P3, P4, and P5 feature maps with downsampling ratios of 8, 16, and 32 respectively.

Owing to differences in fusion paths and integration levels within PAN, feature maps at different scales exhibit distinct semantic characteristics, with richer semantic information generally emerging at lower spatial resolutions. In YOLO-based object detection, prior studies have demonstrated that feature maps of different resolutions are responsible for targets at corresponding scales, and all three scales contribute substantially to overall detection performance.

However, in box-driven pose estimation models, positive sample assignment is primarily determined by the bounding-box localization task, which imposes relatively weak semantic constraints. As a result, feature maps at different scales are generally sufficient for box localization, whereas pose regression is treated as an auxiliary branch whose feature selection is largely influenced by detection-oriented semantics. 
This introduces a task-level semantic misalignment between box-driven optimization and pose-specific regression learning.

Motivated by this observation, we conducted architectural ablation experiments to evaluate the impact of the box branch on pose estimation performance. The results are summarized in \cref{tab:box-driven ablation experiments}.

\begin{table}[!h]
  \centering
  \renewcommand{\arraystretch}{1.4}
  \caption{Semantic ablation experiments on the box-based branch}
  \label{tab:box-driven ablation experiments}
  \begin{tabular}{cccccc}
  \hline
  Box branch & \makecell{Keypoint-driven\\Assignment}  & AP  & $AP_{50}$ & $AP_{75}$ & AR  \\ \hline
  $\checkmark$   & $\times$                   & 50.4 & 79.5 & 54.2 & 57.8 \\
  $\checkmark$   & $\checkmark$                   & 54.7 & 83.2 & 58.8 & 60.2 \\
  $\times$   & $\checkmark$                   & 57.1 & 84.6 & 62.0 & 62.4 \\ \hline
  \end{tabular}
\end{table}

\begin{figure*}[!ht]
  \centering
  \begin{subfigure}{0.89\textwidth}
    \centering
    \includegraphics[width=\textwidth]{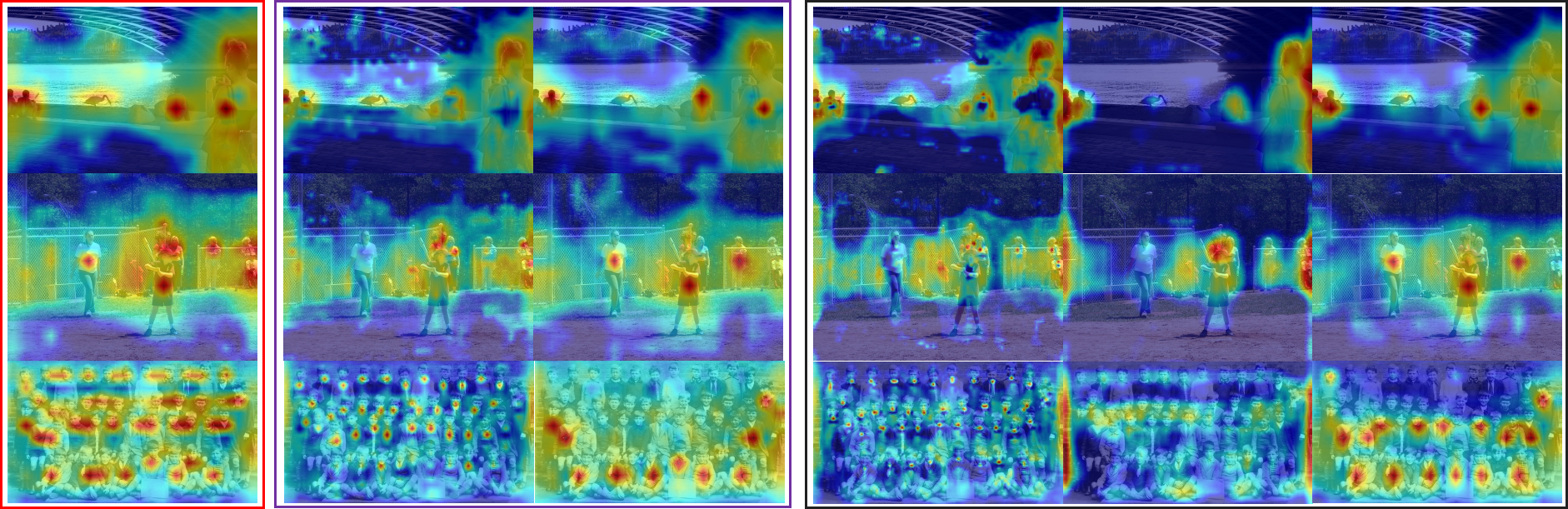}
    \caption{ER-Pose-n}
  \end{subfigure}
  \begin{subfigure}{0.89\textwidth}
    \centering
    \includegraphics[width=\textwidth]{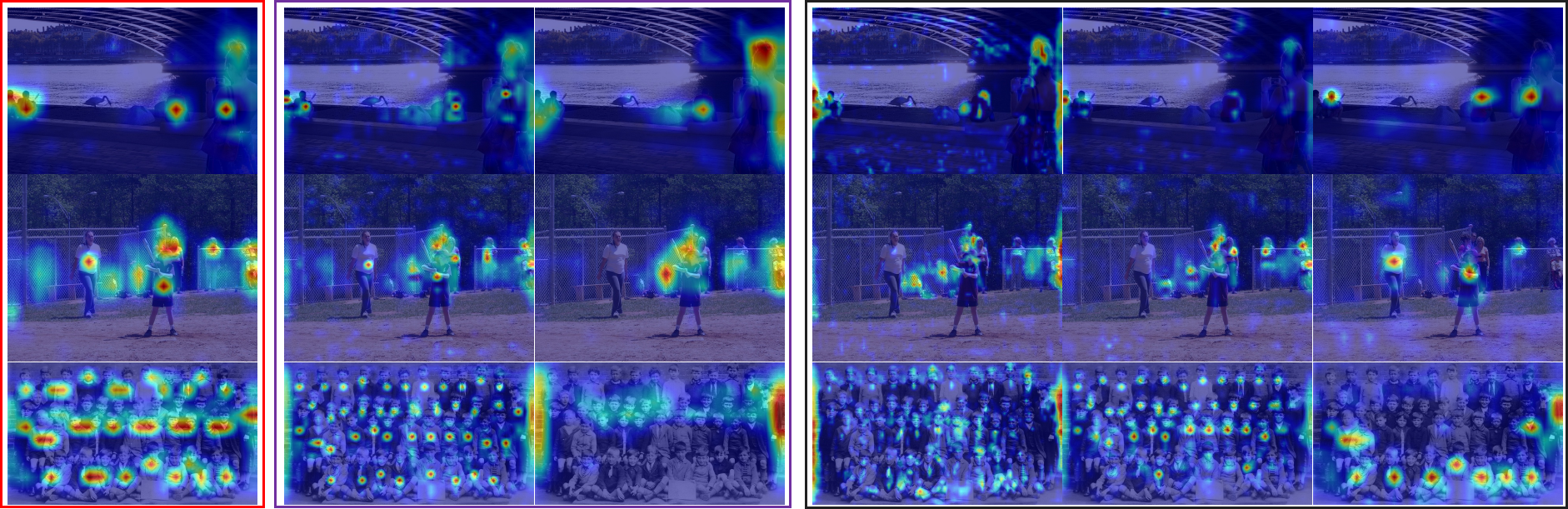}
    \caption{ER-Pose-s}
  \end{subfigure}
    \caption{Gradient heatmaps of confidence prediction branches for models of tiny and small variants.
    From left to right, results correspond to models using one, two, and three feature scales.
    (a) Grad-CAM visualization of ER-Pose-n.
    (b) Grad-CAM visualization of ER-Pose-s.}
  \label{figure:branchs ablation}
\end{figure*}
When the model relies solely on box-driven supervision without pose-centric assignment, it achieves 50.4 mAP. Introducing pose-centric assignment while retaining the box branch improves performance to 54.7 mAP, demonstrating the benefit of aligning sample assignment with pose objectives.

Further removing the box branch while maintaining pose-centric assignment leads to consistent improvements across all evaluation metrics, achieving 57.1 mAP. The steady performance gains across configurations indicate that decoupling pose regression from box-driven supervision enhances overall pose estimation quality and validates the effectiveness of the fully keypoint-driven design.


Furthermore, regression-based human pose estimation can be viewed as a detection problem with substantially higher semantic complexity, involving the regression of $N$ unordered keypoint vectors rather than four ordered bounding-box parameters. Therefore, the semantic compatibility between feature maps of different scales and pose regression deserves further investigation.

To this end, we conducted extensive experiments to analyze the contributions of feature maps at different scales across models of varying capacities. The findings are used to guide the redesign of the prediction head. Specifically, the effects of individual feature scales on prediction accuracy and recall are evaluated, with the results reported in \cref{tab:branch ablation}. 
\begin{table}[!h]
  \centering
  \renewcommand{\arraystretch}{1.4}
  \caption{Branch ablation experiments}
  \label{tab:branch ablation}
  \begin{tabular}{c|c|cccccc}
  \hline
  scale              & features       & AP & $AP_{m}$ & $AP_{l}$ & AR &$AR_{m}$ &$AR_{l}$\\ \hline
  \multirow{3}{*}{n} & \{P5\}         &55.9    &46.7    &\textbf{69.1}    &60.4 &51.2 &\textbf{73.2} \\ \cline{2-8}
                     & \{P4, P5\}     &\textbf{57.1}    &\textbf{49.1}    &68.8    &\textbf{62.4} &\textbf{54.9} &73.0 \\ \cline{2-8} 
                     & \{P3, P4, P5\} &56.4    &47.5    &68.9    &61.6 &53.4 &73.0 \\ \hline
  \multirow{2}{*}{s} & \{P5\}         &64.0    &56.1    &\textbf{75.9}    &68.4 &60.5 &\textbf{79.7} \\ \cline{2-8}
                     & \{P4, P5\}     &64.0    &\textbf{58.9}    &72.1    &68.6 &63.4 &75.9 \\ \cline{2-8} 
                     & \{P3, P4, P5\} &\textbf{65.7}    &58.7    &75.9    &\textbf{70.5} &\textbf{63.9} &79.7 \\ \hline
  \multirow{2}{*}{m} & \{P5\}         &68.2    &61.0    &\textbf{79.2}    &72.5 &65.4 &\textbf{82.7}  \\ \cline{2-8}
                     & \{P4, P5\}     &68.3    &\textbf{64.9}    &76.3    &72.9 &68.5 &78.9  \\ \cline{2-8} 
                     & \{P3, P4, P5\} &\textbf{69.7}    &64.7    &77.4    &\textbf{73.8} &\textbf{69.0} &80.8  \\ \hline
  \end{tabular}
\end{table}


The experimental results indicate that when only the P5 feature map is utilized, its lower spatial resolution favors semantic abstraction for large-scale pose instances, yet leads to relatively limited overall recall.
This suggests that relying solely on high-level semantic features is insufficient for adequately covering densely distributed human instances.

Introducing the P4 and P3 branches significantly improves recall, as higher-resolution feature maps provide richer local structural cues that enhance responsiveness to medium- and small-scale instances.

Notably, the impact of multi-scale features varies across model capacities.
For small and larger models, incorporating P4 slightly compromises performance on large-scale instances while substantially improving medium- and small-scale accuracy.
Further integrating the P3 branch enables the network to leverage fine-grained structural information, resulting in overall performance gains.
These observations suggest that with sufficient model capacity, the network can effectively integrate multi-scale semantic abstraction and spatial detail, achieving coordinated optimization across scales.
In contrast, for the tiny model, limited representational capacity constrains the semantic expressiveness of high-resolution branches, preventing them from forming sufficiently discriminative features and thus limiting performance gains.

To provide qualitative insight, we visualize gradient heatmaps of different feature branches in the tiny and small-scale ER-Pose models under varying branch configurations, as shown in Fig.~\ref{figure:branchs ablation}.

\begin{figure*}[!h]
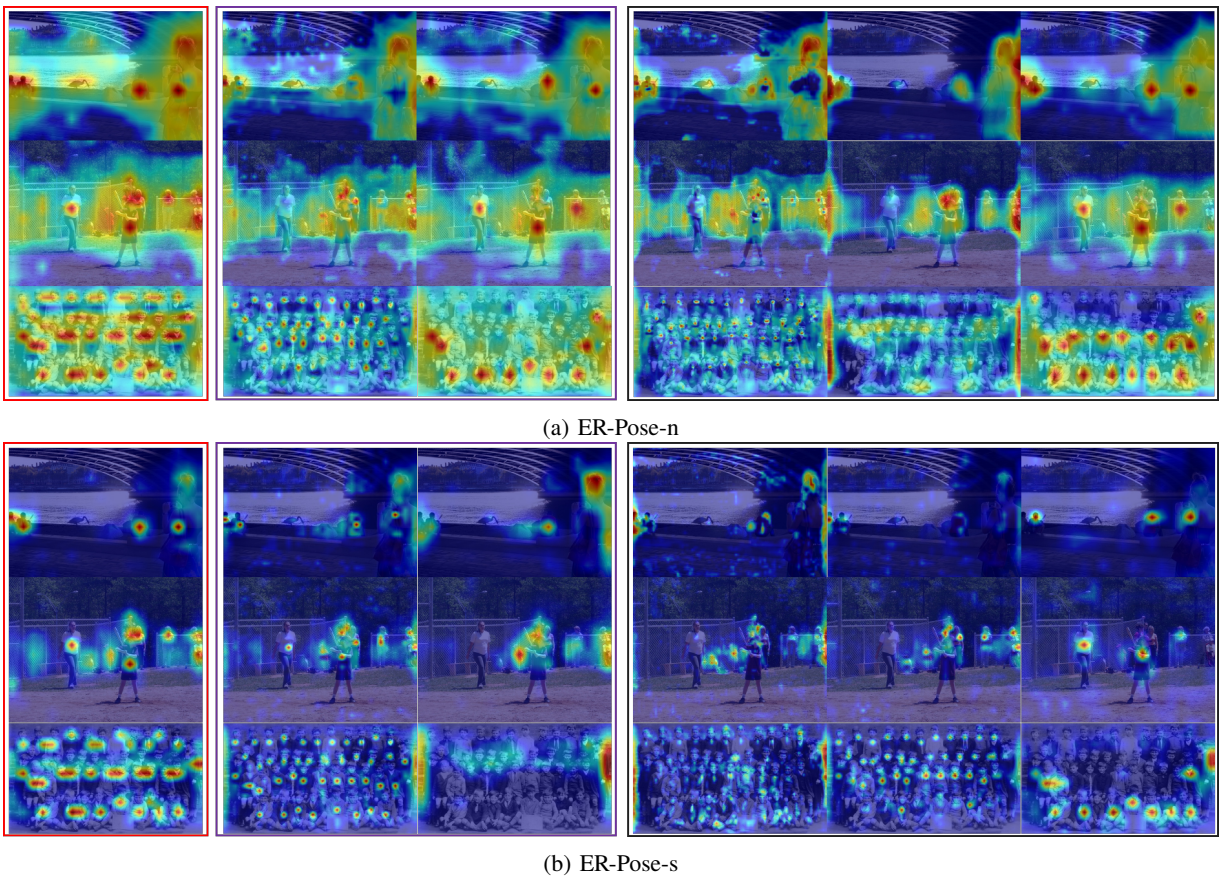

  \centering
  \begin{subfigure}{0.89\textwidth}
    \centering
    \includegraphics[width=\textwidth]{images/ER-POSE-N-Heatmap.png}
    \caption{ER-Pose-n}
  \end{subfigure}
  \begin{subfigure}{0.89\textwidth}
    \centering
    \includegraphics[width=\textwidth]{images/ER-POSE-S-Heatmap.png}
    \caption{ER-Pose-s}
  \end{subfigure}
    \caption{Gradient heatmaps of confidence prediction branches for models of tiny and small variants.
    From left to right, results correspond to models using one, two, and three feature scales.
    (a) Grad-CAM visualization of ER-Pose-n.
    (b) Grad-CAM visualization of ER-Pose-s.}
  \label{figure:branchs ablation}
\end{figure*}

After introducing the P3 branch, models at the small scale and above exhibit more concentrated and coherent gradient distributions, indicating stable and structured feature learning.
In contrast, the tiny model displays sparse and spatially fragmented gradient responses, suggesting unstable optimization dynamics and insufficient semantic abstraction.
Under limited model capacity, high-resolution features fail to contribute meaningful structural cues and instead introduce optimization noise, thereby degrading overall pose estimation performance.
Specifically:
\begin{itemize}
\item When model capacity is sufficient, high-resolution branches complement semantically rich features by providing fine-grained spatial cues, enabling coordinated optimization across object scales;
\item When model capacity is constrained, the network lacks sufficient representational ability to extract discriminative semantics from high-resolution features, causing this branch to act as feature-level interference rather than meaningful structural enhancement.
\end{itemize}





Notably, this observation contrasts with the conventional multi-scale design principles widely adopted in object detection.
In traditional detection frameworks, feature maps at different scales are typically assigned to objects of corresponding sizes—for instance, high-resolution features are primarily used to improve small-object detection \cite{lin2017feature,lin2017focal,redmon2018yolov3}.

In contrast, human pose estimation does not follow a simple scale-specialized division. Instead, effective pose regression relies on the coordinated interaction between structural detail and semantic abstraction across feature hierarchies.
Consequently, multi-scale design strategies directly inherited from object detection do not necessarily transfer to pose estimation tasks.

Based on the above empirical findings, we adopt a capacity–feature matching principle in the head design of ER-Pose. Specifically, lower-level branches are introduced only when the model possesses sufficient representational capacity to effectively leverage high-resolution features, thereby maximizing multi-scale fusion benefits while maintaining optimization stability.
The detailed configurations are summarized in Table~\ref{tab:protocol}.
\begin{table}[!h]
  \centering
  \renewcommand{\arraystretch}{1.4}
  \setlength{\tabcolsep}{12pt}
  \caption{Features maps protocol}
  \begin{tabular}{cc}
  \hline
  scale & features       \\ \hline
  n     & \{P4, P5\}     \\ \hline
  s     & \{P3, P4, P5\} \\ \hline
  m     & \{P3, P4, P5\} \\ \hline
  l     & \{P3, P4, P5\} \\ \hline
  \end{tabular}
  \label{tab:protocol}
\end{table}

\subsection{Dual-Head Assignment Analysis}
The dual-head assignment strategy is adopted for both training and inference. The one-to-many head provides dense supervision during training, whereas the one-to-one head is retained at inference, enabling NMS-free prediction and facilitating decoupling of pose estimation from object detection.

To systematically examine the role and limitations of NMS in single-stage multi-person pose estimation, we conducted experiments from two perspectives: sensitivity to the NMS threshold and comparison of different inference-time post-processing strategies. The corresponding results are presented in \cref{tab: nms-oks vs performence} and \cref{tab:SAH-MAH} respectively. 
ER-Pose adopts OKS-based NMS whereas the Box-driven model employs IoU-based NMS.
All experiments were conducted on a single NVIDIA Tesla V100 GPU and all evaluated models adopt the tiny configuration.

\begin{table}[!h]
  \centering
  \caption{Model performance comparison under different positive sample assignment centers and NMS strategies}
  \label{tab: nms-oks vs performence}
  \renewcommand{\arraystretch}{1.4}
  \begin{tabular}{ccccccc}
    \hline
    \multirow{2}{*}{IoU thr.} & \multicolumn{3}{c}{ER-Pose(OKS-IoU)}      & \multicolumn{3}{c}{Box-driven(IoU)}    \\
                              & AP            & AR            & NMS(ms)   & AP           & AR            & NMS(ms) \\ \hline
    0.4                       & 57.1          & 61.5          &7.434           & 49.9          & 56.3          &0.978         \\
    0.5                       & 57.0          & 61.7          &8.769           & 50.1          & 56.7          &1.085         \\
    0.6              & 56.9          & 61.8          &9.746           & 50.3          & 57.1          &1.034       \\
    0.7                       & 56.6          & 61.9          &12.199          & 50.4          & 57.4          &1.040         \\
    0.8                       & 56.2          & 62.2          &13.791          & 50.4 & 57.8 &0.951         \\ \hline
  \end{tabular}
\end{table}

\begin{table}[!h]
  \centering
  \renewcommand{\arraystretch}{1.4}
    \caption{Ablation results of multi-head structures.
    SAH denotes models trained using single-head assignment, where inference relies exclusively on SAH.
    Models using only MAH adopt OKS-based NMS during inference to suppress redundant predictions.}
  \begin{tabular}{cccccc}
  \hline
  model & AP & $AP_{50}$ & $AP_{75}$ & AR & latency(ms) \\ \hline
  SAH      &53.1     &81.8      &56.7      &58.8     &10.587         \\
  MAH+NMS(OKS)      &56.6     &83.4      &61.4      &60.8     &13.544   \\
  MAH+SAH      &57.1     &84.6      &62.0      &62.4     &10.721         \\ \hline
  \end{tabular}
  \label{tab:SAH-MAH}
\end{table}

The experimental results indicate that although adjusting the OKS-based NMS threshold influences model outputs to some extent, OKS-based NMS consistently achieves lower pose estimation accuracy than the inference scheme based on MAH–SAH multi-head assignment without NMS. Moreover, compared with IoU-based NMS, OKS-based NMS incurs higher inference latency, leading to additional computational overhead.

As shown in Table 4, training with SAH alone results in a noticeable decline in model accuracy. In contrast, the proposed keypoint-driven modeling combined with the multi-head assignment mechanism directly produces final pose predictions at inference time, achieving superior accuracy and efficiency compared with both the NMS-dependent MAH scheme and the SAH-only configuration.

\begin{figure}[!h]
  \centering
  \includegraphics[width=0.45\textwidth]{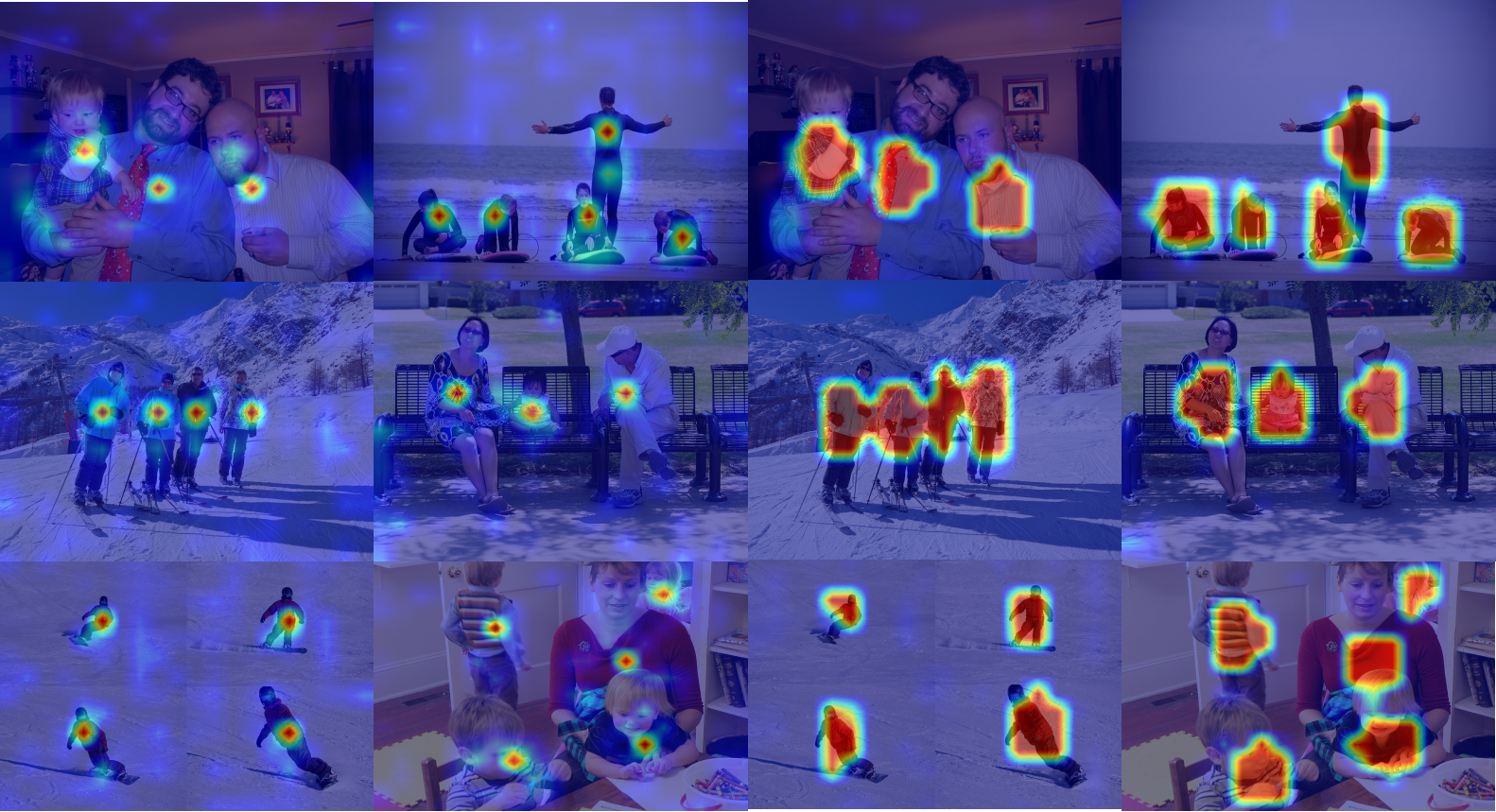}
    \caption{Grad-CAM visualizations of dual-head structures.}
  \label{figure:Dual-head Grad-Cam}
\end{figure}

These results suggest that the dual-head architecture effectively integrates the complementary strengths of both components, making it a favorable design for regression-based pose estimation. To further illustrate the behavioral differences, Grad-CAM visualizations of the single-head and dual-head models are provided in Fig. \ref{figure:Dual-head Grad-Cam}.

Specifically, MAH assigns a broader region around human centers as positive samples during training, whereas SAH concentrates supervision on more compact regions, typically localized to single grid positions, as illustrated in \cref{figure:Dual-head Grad-Cam}.

\begin{figure}[!h]
  \centering
  \includegraphics[width=0.45\textwidth]{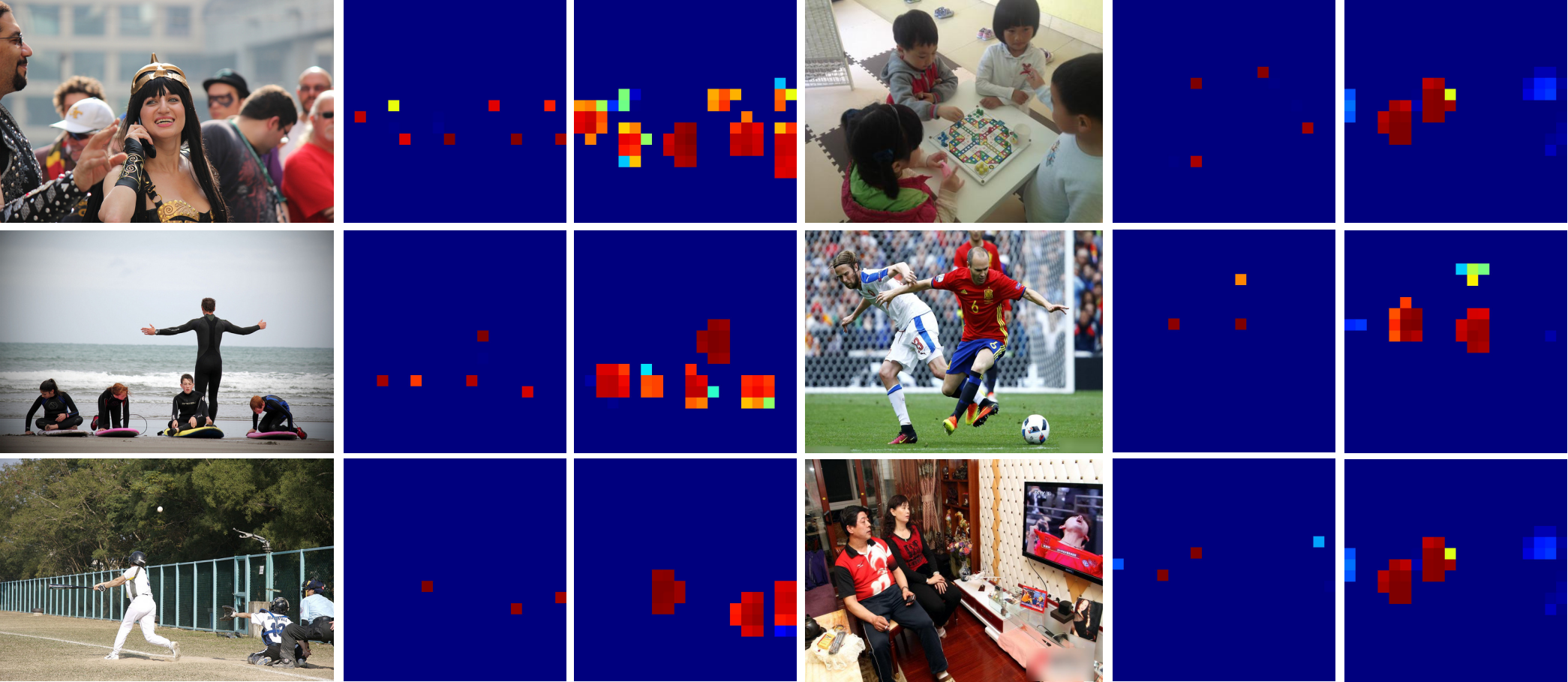}
  \caption{Visualization of human target confidence maps.}
  \label{figure:inference maps}
\end{figure}

To further elucidate the effects of the dual-head strategy, confidence feature maps produced by both branches at inference are visualized. These visualizations provide intuitive evidence of how the proposed mechanism avoids reliance on NMS and improves inference efficiency, as shown in \cref{figure:inference maps}.
SAH produces isolated pixel-level responses, enabling direct localization of the human center.
In contrast, MAH generates dense responses over multiple neighboring pixels,
requiring redundant candidate centers to be filtered in order to identify the final prediction.
Positive sample suppression is therefore implicitly introduced during this selection process.

\subsection{Keypoint-Driven Assignment Metric Evaluation}
\label{sec: assignment error}
To assess the impact of the proposed keypoint-driven assignment metric on training optimization, experiments were conducted under different settings of $\alpha$ and $\beta$, which control the relative weighting between human localization and keypoint estimation during assignment. Performance is evaluated on the val2017 split, with the results summarized in \cref{table:parameters}.
All evaluated models adopt the tiny configuration.
\begin{table}[!h]
  \centering
  \renewcommand{\arraystretch}{1.3}
  \setlength{\tabcolsep}{10pt}
  \caption{Comparison experiments under different $\alpha$ and $\beta$ settings.}
  \label{table:parameters}
  \begin{tabular}{cccccc}
  \hline
  $\alpha$ &$\beta$ & AP  & $AP_{50}$  & $AP_{75}$  & AR \\ 
  \hline
  0.5 &0  &50.7  &78.9  &54.6  &56.7  \\
  0.5 &2  &46.2  &80.6  &46.4  &52.0  \\
  0.5 &4  &54.8  &82.9  &58.8  &60.4  \\
  \textbf{0.5} &\textbf{6} &\textbf{57.1} &\textbf{84.4} &\textbf{62.4} &\textbf{62.2} \\
  0.5 &8  &56.7  &84.4  &61.2  &62.1  \\
  1.0 &0  &48.7  &77.7  &52.2  &54.1  \\
  1.0 &2  &53.0  &81.4  &57.4  &57.7  \\
  1.0 &4  &53.7  &81.9  &58.5  &59.5  \\
  1.0 &6  &54.3  &81.9  &58.8  &60.0  \\
  1.0 &8  &55.5  &83.5  &59.7  &60.7  \\
  \hline
  \end{tabular}
\end{table}

Experimental results indicate that the proposed assignment metric consistently improves human pose estimation performance, with multiple parameter settings yielding stable gains. Based on comprehensive evaluation, $\alpha=0.5$ and $\beta=6$ are selected as the optimal configuration. Compared with strategies that rely solely on human confidence for assignment (i.e., $\alpha=0$), the pose-classification assignment with $\beta=6$ achieves improvements of 6.3 mAP and 5.6 mAP over the baseline settings.

\begin{figure}[!h]
  \centering
  \includegraphics[width=0.48\textwidth]{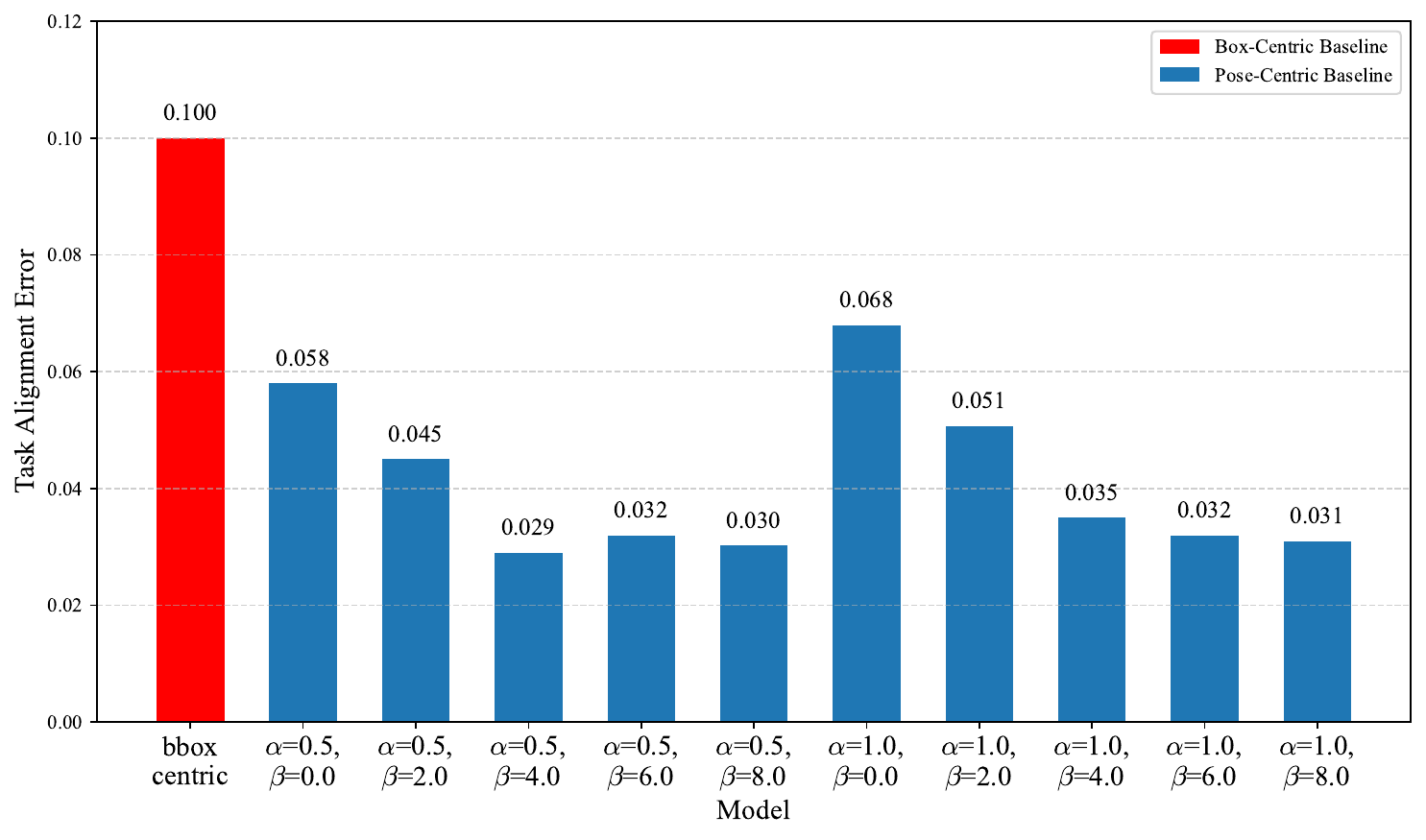}
    \caption{Task alignment deviation under different $\alpha$ and $\beta$ settings.
    Red bars denote alignment deviation of box-driven models, where positive samples are assigned based on box localization.
    Blue bars represent alignment deviation of ER-Pose-n architecture.}
  \label{figure:contribution}
\end{figure}

To provide a quantitative analysis of the optimization effects introduced by the proposed assignment strategy, the average assignment deviation defined in \cref{fun:TAE} was computed. Evaluation was conducted over all images containing human instances in the val2017 split, covering ER-Pose models under different $\alpha$ and $\beta$ configurations as well as the box-driven baseline. The results are illustrated in \cref{figure:contribution}.

The task alignment deviation of box-driven pipelines remains at approximately 0.1, indicating that positive sample locations selected based on confidence and IoU deviate by about 10\% in OKS from the optimal inference locations. Under this paradigm, pose representations tend to be constrained due to their treatment as auxiliary outputs. In contrast, the proposed keypoint-driven assignment metric significantly reduces task alignment deviation, alleviating inconsistencies between training-time positive selection and inference-time optimal predictions, thereby improving overall pose estimation accuracy.

\begin{figure}[!h]
  \centering
  \includegraphics[width=0.4\textwidth]{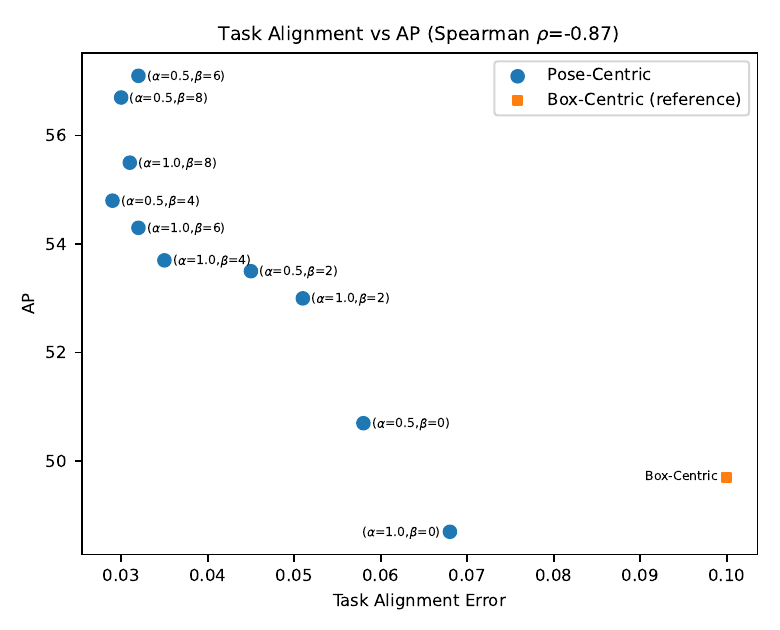}
  \caption{Scatter Plot of TAE vs. AP}
  \label{figure:tae-ap}
\end{figure}

To further illustrate this relationship, we present a scatter plot showing the correlation between TAE and pose accuracy, as shown in \cref{figure:tae-ap}.

As task alignment error increases, model accuracy exhibits a consistent downward trend. The Spearman’s rank correlation coefficient between the two metrics is -0.87, indicating a strong monotonic negative correlation. These results suggest that TAE effectively captures the impact of task-level inconsistency between training and inference on pose estimation performance.

Notably, the box-driven baseline is located in the region characterized by higher alignment error and lower accuracy, suggesting that its performance limitation is closely associated with task inconsistency introduced by the underlying modeling assumptions, rather than solely with model capacity or optimization design.

\begin{figure*}[!h]
  \centering
  \includegraphics[width=0.89\textwidth]{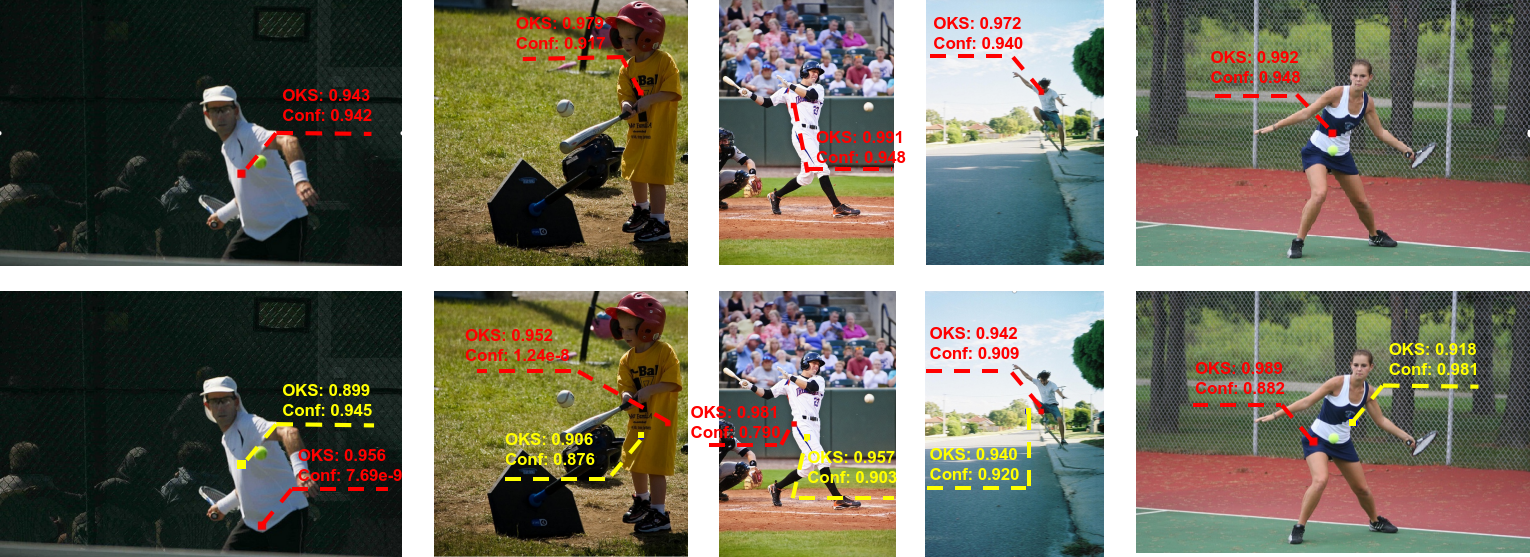}
  \caption{Comparison of task alignment, illustrating effects of keypoint-driven assignment on sample selection.}
  \label{figure:algin}
\end{figure*}
To provide intuitive evidence of task consistency, confidence distributions and their corresponding OKS locations are visualized for selected samples, as shown in \cref{figure:algin}.
In box-driven models, confidence scores are primarily influenced by bounding-box localization quality, leading to inconsistencies between prediction confidence and actual pose quality (OKS).
As shown in several examples, high-OKS predictions may receive lower confidence scores and be suppressed during inference.
In contrast, the proposed keypoint-driven framework aligns confidence estimation with pose quality. Predictions with higher OKS consistently correspond to higher confidence, resulting in more reliable ranking and improved final performance.
This observation provides intuitive evidence for the task misalignment phenomenon and further validates the effectiveness of our assignment reformulation.

\subsection{Loss Function Ablation Study}
To investigate the effect of the proposed SOKS regression loss on training performance, we conducted a comparative study of three loss formulations, all defined on the regression variable $u$: a Laplace-based constraint, a Gaussian-based OKS loss, and the proposed SOKS loss.
The results are reported in \cref{tab:loss}.
\begin{table}[!h]
  \centering
  \renewcommand{\arraystretch}{1.4}
  \caption{Performance evaluation under different loss functions.}
  \label{tab:loss}
  \begin{tabular}{ccccccc}
  \hline
  \multirow{2}{*}{Loss} & \multicolumn{2}{c}{ER-Pose-n} & \multicolumn{2}{c}{ER-Pose-s} & \multicolumn{2}{c}{ER-Pose-m} \\ \cline{2-7} 
                        & AP       & AR       & AP       & AR       & AP       & AR       \\ \hline
    Lap         & 55.7      &60.6       &62.2       &67.2       & 68.9      &73.0           \\ 
    Gau(OKS)    & 56.1      &61.4       &65.3       &70.1       & 69.1      &73.3           \\ 
    SOKS & \textbf{57.1}      &\textbf{61.6}       &\textbf{65.8}       &\textbf{70.2}       & \textbf{69.7}      &\textbf{73.8}       \\ \hline
  \end{tabular}
\end{table}

As shown in the table, SOKS consistently achieves the best performance across all model scales.

Compared with the Laplace-based constraint, SOKS provides stable improvements in both mAP and mAR, indicating that a single-distribution formulation may struggle to simultaneously ensure optimization stability and high-precision regression across different error ranges. 
Relative to the Gaussian-based OKS loss, SOKS further yields consistent gains across all scales. For example, it achieves +1.0 mAP on ER-Pose-n and +0.6 mAP on ER-Pose-m. 

Overall, these results suggest that the proposed piecewise formulation, integrating the complementary properties of Gaussian and Laplace distributions, offers stronger gradient supervision in large-error regions while preserving stable refinement in small-error regimes. This balanced optimization behavior contributes to improved regression quality.

\begin{table*}[!h]
  \centering
  \renewcommand{\arraystretch}{1.4}
  \setlength{\tabcolsep}{6pt}
  \caption{Evaluation on Crowdpose}
  \label{tab:CrowdPose-EMH}
  \begin{tabular}{cccccccccc}
  \hline
  model   &Backbone  &Input size &Params(M) & AP  & $\mathrm{AP_{50}}$ & $\mathrm{AP_{75}}$ & $\mathrm{AP_E}$  & $\mathrm{AP_M}$  & $\mathrm{AP_H}$ \\ \hline
  SimpleBaseline\cite{xiao2018simple} &ResNet-101 &256$\times$192 &34.0 &60.8 &81.4 &65.7 &71.4 &61.2 &51.2\\
  DEKR\cite{xiao2018simple} &HRNet-W32 &512$\times$512 &65.7 &67.3 &86.4 &72.2 &74.6 &68.1 &58.7\\
  CID\cite{wang2022contextual}&HRNet-W32 &512$\times$512 &29.4 &71.2 &89.8 &76.7 &77.9 &71.9 &63.8 \\
  HigherHRNet\cite{cheng2020higherhrnet}&HRNet-W32 &512$\times$512 &63.8 &65.9 &86.4 &70.6 &73.3 &66.5 &57.9\\
  YOLO-Pose-s\cite{mcnally2022rethinking}&CSPDarknet  &640$\times$640 &11.6 & 61.9 &87.6 &67.8 & 69.3 & 62.5 & 53.9  \\
  YOLO-Pose-m\cite{mcnally2022rethinking}&CSPDarknet &640$\times$640 &26.4 & 66.6 &89.4 &73.7 & 73.8 & 67.3 & 58.8  \\
  YOLO-Pose-l\cite{mcnally2022rethinking}&CSPDarknet  &640$\times$640 &44.4 & 69.3 &90.6 &76.2 & 76.1 & 70.2 & 61.5  \\
  KAPAO-s\cite{mcnally2022rethinking}&CSPDarknet &1280$\times$1280 &12.6 & 63.8 &87.7 &69.4 & 72.1 & 64.8 & 53.2 \\
  KAPAO-m\cite{mcnally2022rethinking}&CSPDarknet  &1280$\times$1280 &35.8 & 67.1 &88.8 &73.4 & 75.2 & 68.1 & 56.9 \\
  KAPAO-l\cite{mcnally2022rethinking}&CSPDarknet &1280$\times$1280 &77.0 & 68.9 &89.4 &75.6 & 76.6 & 69.9 & 59.5 \\
  ER-pose-s&CSPDarknet  &640$\times$640 &9.69 & 66.6 &89.4 &72.9 & 73.7 & 67.4 & 58.2 \\ 
  ER-pose-m&CSPDarknet  &640$\times$640 &23.36 & 69.4 &90.1 &75.9 & 76.4 & 69.9 & 61.9 \\ 
  ER-pose-l&CSPDarknet  &640$\times$640 &39.95 & 72.5 &91.2 &79.1 & 78.8 & 73.1 & 65.0 \\ \hline 
  \end{tabular}
\end{table*}

\subsection{Evaluation on CrowdPose}
To evaluate performance under crowded scenarios, we conducted experiments on the CrowdPose dataset. The model is trained on the 12K-image trainval split and evaluated on the 8K-image testval split. All training settings remain consistent with those used for COCO.
\cref{tab:CrowdPose-EMH} presents the comparison of model complexity and performance between ER-Pose and state-of-the-art methods.

Most state-of-the-art methods on CrowdPose rely on heatmap-based representations with spatial smoothing. In contrast, the regression-based ER-Pose achieves competitive performance without requiring inference-time post-processing or optimization heuristics, while maintaining a smaller model size and faster inference speed.

Notably, ER-Pose demonstrates stronger performance on the hard subset, suggesting that regression-based vector representations are more effective than per-keypoint independent heatmap representations in modeling structural relationships among human keypoints.

\begin{figure}[!h]
  \centering
  \begin{subfigure}{0.14\textwidth}
      \centering
      \includegraphics[width=\linewidth]{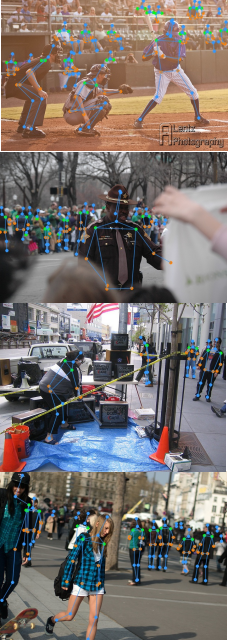}
      \caption{YOLO-Pose-s}
      \label{figure:YOLO-Pose-s}
  \end{subfigure}
  \begin{subfigure}{0.14\textwidth}
      \centering
      \includegraphics[width=\linewidth]{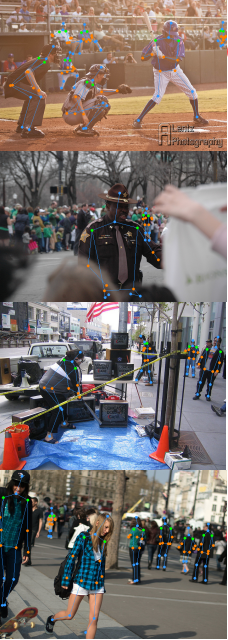}
      \caption{KAPAO-s}
      \label{figure:KAPAO-s}
  \end{subfigure}
  \begin{subfigure}{0.14\textwidth}
    \centering
    \includegraphics[width=\linewidth]{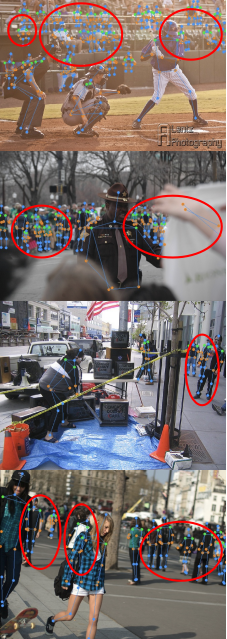}
    \caption{ER-Pose-s}
    \label{figure:ER-Pose-s}
  \end{subfigure}
\caption{Visualization comparison of ER-Pose and advanced regression methods on partial images of CrowdPose.
(a) YOLO-Pose-s test results. (b) KAPAO-s test results. (c) ER-Pose-s test results.
ER-Pose can better localize humans in complex and occluded scenes, and can achieve better accuracy.}
  \label{figure:Visiualization on CrwodPose}
\end{figure}

Qualitative comparisons are provided in \cref{figure:Visiualization on CrwodPose}, where prediction results of three purely regression-based methods—YOLO-Pose-n, KAPAO-s, and ER-Pose-n—are shown for complex crowded scenes.

\subsection{Robustness to Initialization}
To further evaluate the inherent effectiveness of the proposed keypoint-driven modeling framework,
we conducted additional comparative experiments between ER-Pose and its baseline YOLO-Pose under a training-from-scratch setting.
Compared with configurations that rely on pretrained weights, training from scratch more directly reflects the model’s optimization capability and architectural soundness.
The corresponding results on the COCO, CrowdPose, and OCHuman datasets are reported in Table \ref{tab:robust}.
OCHuman is a benchmark designed to evaluate performance in heavily occluded and crowded scenarios, consisting of 4,731 images and over 8,000 human instances.
Following the standard evaluation protocol, we directly evaluate models trained on COCO without additional fine-tuning.
All evaluated models adopt the tiny variant.

\begin{table}[!h]
    \centering
    \renewcommand{\arraystretch}{1.4}
    \caption{Comparison of AP with and without pretraining on COCO, CrowdPose, and OCHuman,
    relative to baseline YOLO-Pose.}
    \label{tab:robust}
    \begin{tabular}{lcccc}
        \toprule
        Setting & Dataset & YOLO-Pose-n & ER-POSE-n & $\Delta AP$ \\
        \midrule
        From scratch & COCO &47.5  &50.7  &\textbf{+3.2}  \\
        Pretrained & COCO &50.4  &57.1  &\textbf{+6.7}  \\
        From scratch & CrowdPose &45.5  &52.9  &\textbf{+7.4}  \\
        Pretrained & CrowdPose &53.4  &58.3  &\textbf{+4.9}  \\
        Pretrained(COCO) & OCHuman &30.7  &37.1  &\textbf{+6.4}  \\
        \bottomrule
    \end{tabular}
\end{table}

\begin{figure*}[!h]
  \centering
  \includegraphics[width=0.89\textwidth]{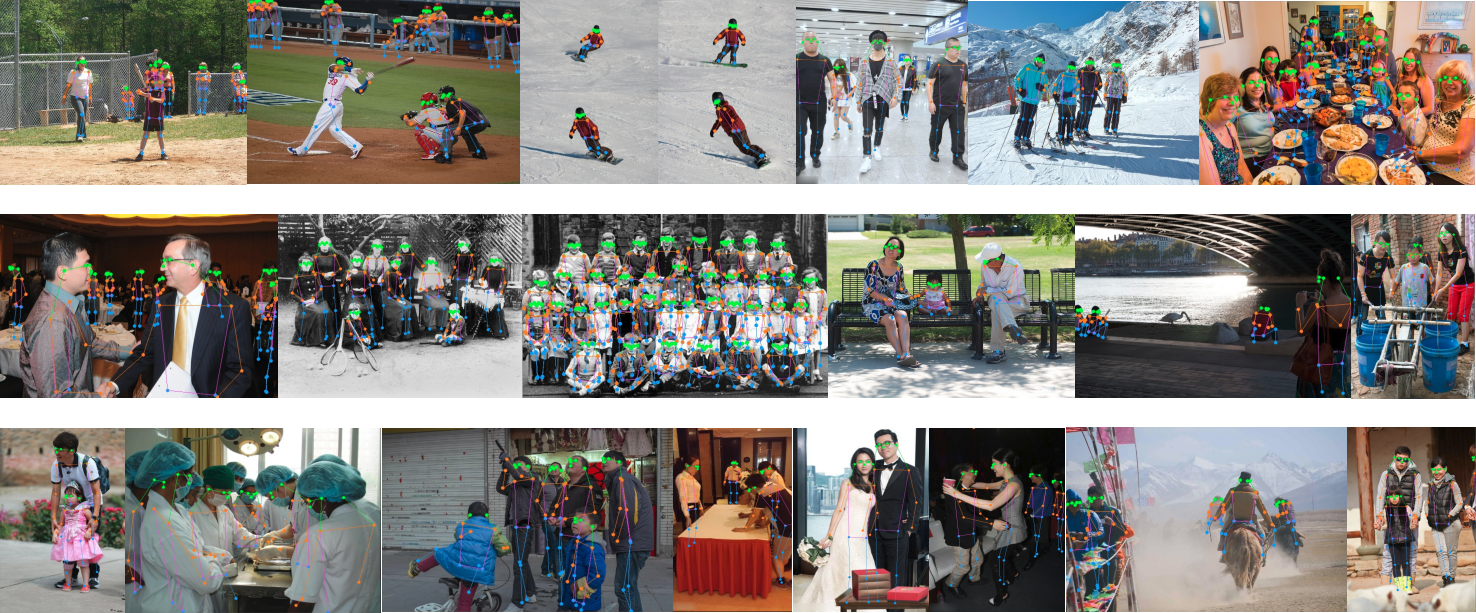}
  \caption{Visualization examples of ER-Pose-n predictions across diverse scenarios.}
  \label{figure:results}
\end{figure*}
Under both pretrained and trained-from-scratch settings, ER-Pose consistently outperforms the baseline YOLO-Pose across all datasets.

When trained from scratch,
ER-Pose achieves gains of +3.2 mAP on COCO and +7.4 mAP on CrowdPose.
These results indicate that, even without pretrained features,
the proposed keypoint-driven modeling and assignment mechanism significantly enhances optimization quality and predictive performance.

Under the pretrained setting, ER-Pose continues to deliver consistent improvements,
achieving mAP gains of +7.7, +4.9, and +6.4 on COCO, CrowdPose, and OCHuman, respectively.
These findings suggest that the performance advantage of the proposed method does not stem from pretrained initialization,
but rather from improvements in architectural design and training mechanisms.

\subsection{Visualization results}
\cref{figure:results} presents some inference results across diverse environmental conditions, including common scenes, crowded scenes, occluded scenes and blurred scenes. Evaluations were conducted using ER-Pose-n. The results indicate that reliable pose estimation quality is maintained even under highly crowded and heavily occluded conditions.
\label{sec:vis}

\section{Conclusion}
This work provides a systematic analysis of the box-driven modeling, training, and inference paradigm widely adopted in single-stage multi-person pose estimation. The analysis reveals a structural task inconsistency: keypoint prediction is tightly coupled with detection objectives, causing pose estimation quality to be constrained by bounding-box localization.

To address this limitation, we introduce ER-Pose, a keypoint-driven regression framework built upon a YOLO-like architecture. By removing the bounding-box branch and directly regressing keypoint offset vectors relative to localization points, the proposed design aligns feature learning and supervision more closely with the semantic requirements of pose estimation. A MAH–SAH dual-head assignment mechanism is further adapted to balance dense training supervision with efficient NMS-free inference. In addition, a keypoint-driven assignment metric is introduced to reduce task misalignment, and a Smooth-OKS (SOKS) loss is proposed to enhance gradient stability in large-error regimes.

Extensive experiments on COCO, CrowdPose, and OCHuman demonstrate that ER-Pose achieves competitive accuracy while maintainng high efficiency and strong robustness under crowded and occluded conditions.

Overall, the results highlight the importance of transitioning from box-driven to keypoint-driven modeling for real-time multi-person pose estimation. Future work will explore extending the proposed formulation to video-based pose estimation, 3D pose estimation, and broader human-centric understanding tasks.

\ifCLASSOPTIONcaptionsoff
  \newpage
\fi

\bibliographystyle{IEEEtran}
\bibliography{ref}

\begingroup
\raggedbottom

\end{document}